# Exploring the law of text geographic information

**Zhenhua Wang, Daiyu Zhang, Ming Ren\*, Guang Xu**

School of Information Resource Management, Renmin University of China, Beijing, China
{zhenhua.wang; zhangdaiyu; renm; 2020000919}@ruc.edu.cn

*Abstract*: Textual geographic information is indispensable and heavily relied upon in practical applications. The absence of clear distribution poses challenges in effectively harnessing geographic information, thereby driving our quest for exploration. We contend that geographic information is influenced by human behavior, cognition, expression, and thought processes, and given our intuitive understanding of natural systems, we hypothesize its conformity to the Gamma distribution. Through rigorous experiments on a diverse range of 24 datasets encompassing different languages and types, we have substantiated this hypothesis, unearthing the underlying regularities governing the dimensions of quantity, length, and distance in geographic information. Furthermore, theoretical analyses and comparisons with Gaussian distributions and Zipf's law have refuted the contingency of these laws. Significantly, we have estimated the upper bounds of human utilization of geographic information, pointing towards the existence of uncharted territories. Also, we provide guidance in geographic information extraction. Hope we peer its true countenance uncovering the veil of geographic information.

## 1. INTRODUCTION

Geographic information, intricately woven into textual data through toponyms, place names and locational references, holds invaluable spatial insights that shape our understanding of the geographic realm (Purves et al., 2018; Hu & Adams, 2021; Hu et al., 2022c). The extraction of geographic information from text not only engenders profound advantages within scientific domains like social linguistics and spatial humanities (Melo & Martins, 2017; Stock, 2018; Purves et al., 2019) but also imparts the prowess to confront intricate challenges and make judicious decisions in practical contexts (Stock et al., 2020; Hu et al., 2022). Unleashing the potential for geographic spatial analysis and informed decision-making (Kersten & Klan, 2020), it unveils veiled spatial patterns, discerns intricate interdependencies, and bequeaths actionable intelligence pertinent to urban planning (Milusheva et al., 2021), public health (Allen et al., 2017), disaster surveillance (Scalia et al., 2022), etc. (Arulanandam et al., 2014; Hu & Wang,

22020). By promptly responding to post-disaster rescue entreaties and resource requisitions (Resch et al., 2018), unravelling the nexus between crisis response and masses-social media interplay (Kersten et al., 2019), capturing the nuanced experiential tapestry of travel destinations (Haris & Gan, 2017), and delving into the intricacies of urban emotion dynamics (Resch et al., 2014), a mosaic of insights emerges.

Central to this pursuit is the abstraction of geographic information into geographic entities, providing a generalized and comprehensive approach for extraction (Wang et al., 2020; Hu & Adams, 2021; Hu et al., 2022a; Hu et al., 2022b), involving developing sophisticated machine learning and natural language processing techniques to automatically identify and classify geographic entities within text (Hu et al., 2022c; Kersten et al., 2022).

While advancements in geographic entity extraction have flourished across scientific disciplines, a significant gap remains unaddressed concerning the distribution patterns of geographic entity. These underlying patterns, containing essential logic, determine the boundaries of geographic entity processing and prompt the need to delve deeper into their exploration. By bridging this gap, we can elevate the extraction (inference, recognition, parsing, and matching etc.) of textual geographic entity to unprecedented heights, unravelling new horizons and dynamics for its utilization.

In this paper, the distribution patterns of geographic entities are discerned primarily from three dimensions. The first dimension revolves around quantity, with its well-known frequencies and ranks (*f-r*), particularly emphasized in the work of Zipf's law (Piantadosi, 2014; Linders & Louwerse, 2023). Note that quantity holds a certain equivalence to frequency, there is only one law: *f-r*, which is one of the underlying reasons behind Zipf's law delving solely into this particular facet within natural language. Hence, here, our exploration primarily revolves around the quantity *f-r*. Our idea extends into the second dimension - length. It being regarded as a carrier of semantics, reflects a profound quest for expressive efficiency (Drożdż et al., 2016; Zörnig & Berg, 2023). Beyond its theoretical implications, delving into the length confers tangible benefits in practice, one is that propels the likelihood of complete entity extraction, especially when dealing with intricately nested entities. An intriguing aspect lies in its attribute with own value, which facilitates the exploration of length values themselves, thereby enabling a comprehensive determination of the laws governing length. Thus, this dimension can unfold with the investigation of three distribution patterns: frequency-rank, frequency-length, and length-rank. The third dimension pertains to the inter-geographic entity distance, offering insights into the public's usage patterns concerning geographic entities and the efforts invested in communication (Stanisz et al., 2023). Unraveling its laws carries the potential for groundbreaking technological applications, one prospect is to enhance the likelihood of unearthing unknown entities. Similarly, here there are mainly three distribution patterns demand comprehensive characterization: frequency-rank, frequency-distance, and distance-rank.

Methodically, this paper embarks on the exploration of seven distribution patterns governing geographic entities. Drawing a parallel between the limited cognitive resources of individuals and the observable manifestations in nature and life, we postulate that these patterns tend to follow gamma distributions, each with its distinct set of parameters. Substantiating our hypotheses through comprehensive analyses on 24 diverse datasets, we ascertain the specific parameters, thereby revealing these laws. Of particular significance is our estimation of the upper boundaries encompassing the utilization of geographic entities by individuals, unveiling a profound understanding of human cognition. The juxtaposition of our findings against Gaussian distribution and Zipf's law further accentuates the uniqueness of our research endeavor. Indeed, our study holds the potential to propel the field of Geographic Information Science towards a new level, enriching our understanding of the intricate interplay between geographic information and human cognition. Our contributions are as follows.

1. We present a groundbreaking revelation by unveiling, for the first time, the distribution patterns governing geographic entities. This achievement significantly bridges an existing gap in our understanding of geographic information.
2. A comprehensive exploration leads to the unveiling of a total of seven distinct laws governing geographic entities, considering their quantity, length, and distance. This analysis provides a framework for comprehending the intricacies of geographic information.
3. Geographic entities adhere to different forms of gamma distribution, deviating from the Gaussian distribution, as well as the Zipf's law presented by words.
4. Estimation of the upper boundary regarding the utilization of geographic entities, beyond its scientific significance, also holds great practical value, e.g. powering the extraction of geographic information.

The remainder of the paper is organized as follows. Section 2 introduces the related work on pursuit of deterministic laws and Gamma distribution. Section 3 outlines the hypothesis concerning text-based geographic entity and details the data and method used. The outcomes and key findings pertaining to the hypothesis are presented in Section 4. Section 5 is the discussions surrounding the obtained results. Section 6 offers the implications behind our discovery, and Section 7 concludes the paper.

## 2. RELATED WORK

### 2.1. Pursuit of deterministic laws

As the world confronts uncertainty, the pursuit of deterministic laws gains paramount importance, driving scientific exploration. Physics, throughout history, has revolved around deterministic mathematical laws, unveiling the secrets of planetary trajectories, electromagnetic interactions, and laws of relativity etc. (Longair, 2003). Meanwhile, in the nineteenth century, notable figures like Maxwell and Boltzmann made groundbreaking discoveries by delving into the realm of probabilistic or statistical laws when investigating the motion of particles. Among these remarkable findings was Planck's radiation law (Tsallis et al., 1995), which can be considered a manifestation of the Gamma distribution. The profound realization of Planck, and other luminaries like Einstein was their introduction of probability as a way to decipher the mechanics governing the fundamental constituents of matter. This insight has emerged as one of the most triumphant intellectual pursuits in human history, yielding remarkable results. While the aforementioned examples of probabilistic laws hold true under the assumption of an "ideal case" with no interactions between constituents, the presence of interactions, at least with the surroundings, becomes necessary to establish a state of equilibrium (Landau and Lifshitz, 2013). In essence, the overarching objective of statistical physics is to comprehend how macroscopic behavior arises from microscopic interactions, whether in equilibrium or out of equilibrium. By exploring this interplay between the microscopic and macroscopic realms, statistical physics unveils the fascinating mechanisms that govern the emergence of observable phenomena at various scales.

The exploration of deterministic laws extends far beyond the realm of physics. It has transcended numerous fields, which is not just Mendel's laws of genetics that explore life's remarkable diversity, the laws of neural connectivity and plasticity that illuminate the complex workings of the human mind, and Allee principle of aggregation that reveals ecological population structure. Also, from the Matthew Effect that governs economics dynamics to the Zipf's law that characterizes natural language. Our quest for certainty has yielded invaluable insights, shaping our comprehension of the world around us.

When it comes to textual geographic information, the lack of well-defined laws hampers the ceiling of extraction, matching, reconstructing, and parsing. To fill this critical gap, our research endeavors to uncover hidden patterns within textual geographic information, paving the way for enhanced understanding and utilization of this valuable data source.

## 2.2. Gamma distribution

If the density function of the random variable *X* satisfies Eq.1,

$$f(x, \alpha, \beta) = \lambda e^{-\beta x} x^{\alpha-1}, x \geq 0 \quad (1)$$

then *X* is said to follow a Gamma distribution, denoted as *X* ~ *Gamma*(α, β) (Stacy, 1962). Where, Γ(α) represents the Gamma function, λ is $\beta^\alpha / \Gamma(\alpha)$, α and β are the shape and scale parameters, respectively. α plays a crucial role in determining the shape of the distribution. When α is relatively small, the distribution is more heavily skewed to the right. As α increases, the distribution becomes more symmetric and approaches a normal distribution when α is large. β is in determining the spread or concentration of the distribution. It influences the average or expected value of the random variable and affects the shape of the distribution curve. A smaller value of β corresponds to a more concentrated distribution, where the probability mass is concentrated around the mean. In other words, the random variable has less variability and is more likely to take values close to the expected value. As β increases, the distribution becomes more spread out, and the variability of the random variable increases. The Gamma distribution is often used to model positive-valued variables that exhibit right-skewness, where, the skewness is a measure of the asymmetry of a probability distribution, indicating whether the data is concentrated more towards the left or the right side of the distribution. This means that there is a higher probability of observing smaller values, while larger values occur less frequently but can extend to very large values.

A plethora of phenomena in the nature and life exhibit indications of Gamma distribution. Spanning from simulating the lifetimes and decay of cosmic stars and radioactive elements, to the polygonal networks (Li et al., 2021) and turbulence (Orsi et al., 2021) in natural systems, to the population growth dynamics within ecosystems, to the infectious diseases (Vazquez, 2021) and digestion (Da et al., 2022) in biological mechanisms, to the development (Susko et al., 2003), structure (Meyer & Wilke, 2015), and evolution (Mayrose et al., 2005; Le et al., 2012) of proteins, and even to the growth, decay, and renewal processes within cells (Till et al., 1964; Traub et al., 1998).

The utilization of quantity, involvement of length, and selection faced with geographic information provide glimpses into the cognitive processes and intellectual deliberations of human beings. If we peek at the signs of Gamma distribution from it, it may help to decipher the truth of the geographic world.

## 3. METHOD

We first propose the hypotheses, then describe the collected data, and finally introduce the approach used.

### 3.1. Hypothesis

The distribution patterns of geographic entity are subjected to human cognition and behavior. As cognitive resources are inherently finite (Wang, 2021), the efficient transmission of anticipated geographic entity within the confines of certain capacity becomes a pursuit that demands our attention. Thus, we find ourselves compelled, often on a subconscious level, to engage in the reasonable allocation of outputs for conveying geographic information.

The quantity of geographic entity evidently holds great significance. In our search for optimal communication, we endeavor to minimize the cognitive burden by employing a least set of geographic elements. Consequently, certain pieces of geographic entities become focal points, concentrated and frequently employed to address the demands of everyday discourse. Meanwhile, other geographic entities, which encompass a broader range of possibilities, assume a role of sparse but richness, serving as supplementary components that embellish and augment audiences' understanding of the broader geographic context. This phenomenon is analogous to the distribution of resource loads, resonating with the concentration of top-level organisms within ecological food chains and the broader dispersal of organisms at lower trophic levels (Worm et al., 2002). Such pattern is often elegantly captured by Gamma distribution. Hence, we posit the fundamental hypothesis 1, which concerns frequency and its rank.

**H1: The distribution of geographic entity quantities conforms to the Gamma distribution.**

In addition, the length of geographic entity transcends mere text considerations, embodying the human communication strategies (Drożdż et al., 2016; Zörnig & Berg, 2023). The choice to employ geographic entity with short length stems from the recognition that brevity enhances efficiency in conveying, yet inevitably the meaning or content may be limited. Conversely, longer geographic entity demands a heightened cognitive effort from the audience to fully comprehend them (Barton et al., 2014). Moreover, the presentation of lengthy geographic entity imposes more cognitive load on the speaker, the load yet can be alleviated through the segmentation of entity into multiple succinct ones (Barton et al., 2014). Hence, the frequency of employing shorter geographic entity could surpass that of their lengthier counterparts, and certain lengths of geographic entity are frequently used, which unveil an intriguing skewed pattern that can be shaped by allowing Gamma distributions with multi styles. Formally, we posit the fundamental hypothesis 2, and there are three laws worth determining: frequency-rank, frequency-length, and length-rank.

**H2: Gamma distribution is a compelling candidate for elucidating the geographic entity length.**

The last concern is the inter-geographic entity distance. The distance encapsulates the number of characters separating adjacent entities, engendering a delicate interplay between proximity and comprehensibility (Stanisz et al., 2023). The artful manipulation of distance influences the contextual cohesiveness and communicator-friendliness of the conveyed meaning. Shorter distances foster a tighter integration of ideas, ensuring a flow of meaning. However, in certain scenarios, longer distances are inevitable, even if they require the audience to invest more effort to prevent forgetting. Thus, it becomes paramount to strike a balance between communication efficiency and the depth of information conveyed. This balance can be presented in a more intuitive form: "*When each emergence of geographic entity is a success, when will subsequent successes arrive?*" Just as individuals await their turn, the arrival of the masses carries a potential timing. This synchronization also echoes the underlying principles that shape radiation patterns, showing the consistency between the dynamics of geographic entity and Gamma distribution. Hence, we postulate the fundamental hypothesis 3, and there are three laws: frequency-rank, frequency-distance, and distance-rank.

**H3: The inter-geographic entity distance follows the Gamma distribution.**

The hypotheses put forth aim to unravel the intricate tapestry that governs geographic entity, transcending its mere surface manifestations. Moreover, they offer an opportunity to break the boundaries of knowledge. Ultimately, from the traceability of statistical physics to the landscape of geographic entity, from the vast expanse of celestial bodies to the minute units of text, and from the nature to the humanity itself, it is through this pursuit that we gain clarity and equip ourselves to confront the dynamics embedded within geographic information.

### 3.2. Data description

We collect 24 text datasets from various languages and domains, which serves as a solid support for our research, see Table 1 listed in alphabetical order. Where the geographic information is represented in the form of entities, which are manually annotated, thereby ensuring their accuracy and reliability. These datasets can be accessed from https://tianchi.aliyun.com/.

Table 1: Dataset description.

| Dataset | Domain | Amount | Language | Dataset | Domain | Amount | Language |
|---|---|---|---|---|---|---|---|
| Conll2003 | News | 322159 | English | Ren1998 | News | 1336062 | Chinese |
| De | Wikipedia | 2928056 | German | Ren2014 | News | 23927270 | Chinese |
| En | Wikipedia | 3732302 | English | Restaurant | Restaurant | 93960 | English |
| Es | Wikipedia | 4433036 | Dutch | Resume | Resume | 71352 | Chinese |
| Fned | Military affairs | 1139436 | Chinese | Ru | Wikipedia | 1485750 | Russian |
| Fr | Wikipedia | 4488191 | French | Weibo | Social media | 104887 | Chinese |
| It | Wikipedia | 4854607 | Italian | Wiki | Wikipedia | 87831 | English |
| Literature | Literature | 1279782 | Chinese | Wikn | Wikipedia | 9875216 | English |
| MSRA | News | 2393207 | Chinese | Winer | Wikipedia | 2803429 | English |
| Nl | Wikipedia | 3219330 | Spanish | Wnut16 | Social media | 131880 | English |
| Pl | Wikipedia | 3177942 | Polish | Wnut17 | Social media | 813268 | English |
| Pt | Wikipedia | 4063915 | Portuguese | Zh | Wikipedia | 5994113 | Chinese |

Table 2 presents a statistical analysis of the quantity distribution of geographic entity. The findings reveal the prevalence and frequency of occurrence. It is evident that the vast majority of geographic entities appears only once, with an astonishing proportion exceeding 80% in *Weibo* and *Wnut17* datasets. Furthermore, as the frequency of occurrence increases, the number of corresponding geographic entities decreases. Notably, the observations indicate that the proportion of geographic entity appearing five times hovers around a mere 2%. And in 23 out of the 24 datasets, the proportion of geographic entity appearing ten times is consistently below 1%. These results unveil an intuitive trend characterized by a gradual decline in the occurrence frequency. The data suggest that a select few geographic entities attain popularity, while the vast majority remain infrequently used and uncommon. This can shed light on cognitive behaviors exhibited by humans, such as the principle of minimal effort (Zipf, 2016). It reflects the tendency to employ a minimal set of geographic entities during communication, thereby optimizing efficiency.

Table 2: Statistics (%) of geographic entity quantity.

| Dataset / Quantity | 1 | 2 | 3 | 4 | 5 | 6 | 7 | 8 | 9 | 10 | 11-20 | 21-30 | 31+ |
|---|---|---|---|---|---|---|---|---|---|---|---|---|---|
| Conll2003 | 50.87 | 14.4 | 6.85 | 5.11 | 3.89 | 2.96 | 2.32 | 1.45 | 1.28 | 0.81 | 4.59 | 1.92 | 3.55 |
| De | 68.2 | 14.67 | 5.68 | 2.71 | 1.8 | 1.05 | 0.83 | 0.49 | 0.49 | 0.3 | 1.75 | 0.57 | 1.46 |
| En | 71.06 | 13.07 | 4.69 | 2.48 | 1.53 | 0.88 | 0.73 | 0.46 | 0.36 | 0.43 | 1.64 | 0.58 | 2.09 |
| Es | 62.96 | 14.96 | 6.33 | 3.24 | 2.15 | 1.35 | 1.01 | 0.82 | 0.46 | 0.46 | 2.56 | 0.97 | 2.73 |
| Fned | 57.45 | 16.17 | 6.59 | 4.49 | 2.68 | 1.67 | 1.67 | 1.04 | 1.02 | 0.72 | 3.47 | 1.19 | 1.84 |
| Fr | 66.67 | 13.9 | 5.53 | 2.97 | 1.85 | 1.23 | 0.83 | 0.82 | 0.63 | 0.47 | 2.26 | 0.81 | 2.03 |
| It | 64.21 | 14.83 | 6.06 | 3.37 | 2.01 | 1.24 | 0.96 | 0.83 | 0.7 | 0.47 | 2.21 | 0.86 | 2.25 |
| Literature | 72.67 | 12.22 | 4.51 | 2.4 | 1.55 | 1.13 | 0.88 | 0.71 | 0.47 | 0.29 | 1.88 | 0.6 | 0.69 |
| MSRA | 56.98 | 14.14 | 6.96 | 3.94 | 2.86 | 2.01 | 1.12 | 0.94 | 0.89 | 0.78 | 3.32 | 1.46 | 4.6 |
| Nl | 61.3 | 15.59 | 6.59 | 3.82 | 2.31 | 1.64 | 0.99 | 0.9 | 0.8 | 0.61 | 2.4 | 0.95 | 2.1 |
| Pl | 64.74 | 14.92 | 6.08 | 3.26 | 2.02 | 1.34 | 1.03 | 0.71 | 0.61 | 0.55 | 2.41 | 0.77 | 1.56 |
| Pt | 46.87 | 17.19 | 10.62 | 8.05 | 5.38 | 3.56 | 1.89 | 1.09 | 0.73 | 0.47 | 1.89 | 0.61 | 1.65 |
| Ren1998 | 59.05 | 14.43 | 6.69 | 3.33 | 2.56 | 1.5 | 0.96 | 0.96 | 0.74 | 0.61 | 3.28 | 2.01 | 3.88 |
| Ren2014 | 46.09 | 15.7 | 7.33 | 4.91 | 3.29 | 2.61 | 1.8 | 1.43 | 1.24 | 1.19 | 6.01 | 2.6 | 5.8 |
| Restaurant | 77.4 | 9.82 | 4.26 | 2.36 | 0.99 | 0.68 | 0.38 | 0.3 | 0.3 | 0.46 | 1.38 | 0.39 | 1.28 |
| Resume | 58.33 | 13.89 | 6.94 | 1.39 | 2.78 | 4.17 | 4.17 | 1.39 | 1.39 | 0 | 4.17 | 1.39 | 0 |
| Ru | 66.79 | 13.95 | 5.72 | 3.02 | 1.76 | 1.35 | 0.97 | 0.6 | 0.65 | 0.54 | 2.44 | 0.8 | 1.41 |
| Weibo | 81.2 | 15.38 | 2.56 | 0 | 0 | 0 | 0.85 | 0 | 0 | 0 | 0 | 0 | 0.01 |
| Wiki | 67.38 | 18.3 | 6.28 | 2.96 | 1.51 | 0.66 | 0.43 | 0.38 | 0.27 | 0.25 | 0.56 | 0.36 | 0.66 |
| Wikn | 73.81 | 12.82 | 5.04 | 2.67 | 1.52 | 0.93 | 0.69 | 0.44 | 0.34 | 0.25 | 0.91 | 0.19 | 0.39 |
| Winer | 74.34 | 11.81 | 4.19 | 2.3 | 1.52 | 0.83 | 0.67 | 0.46 | 0.48 | 0.32 | 1.35 | 0.49 | 1.24 |
| Wnut16 | 74.15 | 13.29 | 4.14 | 2.36 | 1.18 | 1.18 | 0.89 | 0.3 | 0.15 | 0.3 | 1.78 | 0.15 | 0.13 |
| Wnut17 | 84.69 | 9.61 | 3.75 | 0.81 | 0.33 | 0.49 | 0.16 | 0 | 0 | 0.16 | 0 | 0 | 0 |
| Zh | 69.86 | 13.6 | 5.69 | 2.79 | 1.51 | 1.09 | 0.77 | 0.57 | 0.42 | 0.37 | 1.66 | 0.5 | 1.17 |

Table 3 analyzes the length distribution of geographic entity. Variations of geographic entities emerge prominently across various datasets, including in *MSRA*, *Fned*, *Ren1998*, *Resume*, and *Literature*, the lengths of geographic entity tend to range from 2 to 4. This pattern aligns with the well-known expressive characteristics of the Chinese language, which often emphasizes succinct forms of communication. Lengthy geographic entities, and even extended sentences, are typically fragmented into multiple subsets for more concise articulation. Remarkably, this observation finds further support in *Ren2014* and *Weibo* datasets. In *Wnut16* and *Conll2003*, geographic entity length tends to cluster around 6 to 9. We believe this is due to the fact that English social media and news often involve commonly encountered geographic entities, which is typically expressed in not too long lengths, aligning with everyday communication norms. *Wnut17* also indirectly reflects this aspect. Turning our attention to datasets like *En*, *Winer*, and *Wiki*, a distinct tendency towards lengths ranging from 11 to 20 becomes apparent. This phenomenon can be attributed to the abundance of proper toponyms and specialized terms prevalent in the English Wikipedia corpus. The inclusion of such terminology inevitably leads to longer lengths.

Overall, a common thread unifies these observations—the datasets exhibit similar patterns characterized by sparse occurrences at both ends (as evident from the proportions of lengths 1 and 31+), with dense concentrations in the middle. This pattern signifies a delicate trade-off wherein employing shorter geographic entity in communication enhances efficiency, albeit with a relatively limited depth of conveyed meaning. Conversely, longer geographic entity poses challenges in effectively expressing and communicating its complexities within the confines of typical communication modes.

Table 3: Statistics (%) of geographic entity length.

| Dataset / Length | 1 | 2 | 3 | 4 | 5 | 6 | 7 | 8 | 9 | 10 | 11-20 | 21-30 | 31+ |
|---|---|---|---|---|---|---|---|---|---|---|---|---|---|
| Conll2003 | 0 | 1.57 | 1.34 | 5.4 | 8.83 | 14.23 | 13.94 | 12.89 | 10.86 | 8.19 | 20.38 | 1.98 | 0.39 |
| De | 0 | 0.1 | 0.65 | 3.18 | 6.58 | 9.13 | 9.73 | 9.36 | 9.49 | 8.06 | 37.48 | 5.52 | 0.72 |
| En | 0.01 | 0.08 | 0.43 | 2.3 | 5.55 | 8.08 | 9.28 | 8.48 | 8.32 | 7.39 | 42.82 | 6.06 | 1.2 |
| Es | 0 | 0.1 | 0.55 | 3.02 | 7 | 9.81 | 11.1 | 9.92 | 8.41 | 6.64 | 33.81 | 8.18 | 1.46 |
| Fned | 8.01 | 50.62 | 24.22 | 11.13 | 2.8 | 1.34 | 0.81 | 0.39 | 0.28 | 0.12 | 0.28 | 0 | 0 |
| Fr | 0 | 0.08 | 0.69 | 3.09 | 6.08 | 8.98 | 9.19 | 8.72 | 7.62 | 6.39 | 37.57 | 9.26 | 2.33 |
| It | 0 | 0.27 | 0.58 | 2.77 | 6.06 | 8.56 | 9.88 | 8.56 | 7.69 | 5.93 | 37.05 | 9.49 | 3.16 |
| Literature | 3.47 | 28.8 | 28.98 | 20.83 | 7.38 | 4.94 | 2.18 | 1.28 | 0.78 | 0.35 | 0.94 | 0.06 | 0.01 |
| MSRA | 3.57 | 22.22 | 37.16 | 15.24 | 8.67 | 4.62 | 3.78 | 2.17 | 1.14 | 0.62 | 0.79 | 0.02 | 0 |
| Nl | 0.01 | 0.12 | 0.95 | 3.98 | 7.57 | 10.12 | 10.74 | 10.1 | 9.51 | 8.41 | 33.51 | 4.49 | 0.49 |
| Pl | 0 | 0.06 | 0.7 | 2.78 | 6.68 | 10.4 | 12.07 | 11.43 | 9.56 | 6.8 | 31.96 | 5.89 | 1.67 |
| Pt | 0 | 0.1 | 0.46 | 2.33 | 6.04 | 9.43 | 10.91 | 10.07 | 8.75 | 6.39 | 36.01 | 7.47 | 2.04 |
| Ren1998 | 4.29 | 24.57 | 37.78 | 14.08 | 7.68 | 4.29 | 3.23 | 1.95 | 0.9 | 0.51 | 0.72 | 0 | 0 |
| Ren2014 | 0 | 13.66 | 25.84 | 20.41 | 15.11 | 14.18 | 3.51 | 2.81 | 2.85 | 0.54 | 1.05 | 0.02 | 0.02 |
| Restaurant | 0 | 0.38 | 0.46 | 1.9 | 1.52 | 3.96 | 5.33 | 5.02 | 6.09 | 6.09 | 52.74 | 13.48 | 3.03 |
| Resume | 0 | 34.72 | 40.28 | 15.28 | 6.94 | 1.39 | 0 | 1.39 | 0 | 0 | 0 | 0 | 0 |
| Ru | 0 | 0.2 | 1.47 | 4.69 | 9.25 | 12.46 | 12.53 | 10.11 | 7.42 | 5.12 | 28.34 | 6.62 | 1.79 |
| Weibo | 2.56 | 41.03 | 29.06 | 15.38 | 6.84 | 4.27 | 0.85 | 0 | 0 | 0 | 0 | 0 | 0.01 |
| Wiki | 0.05 | 0.23 | 0.75 | 3 | 5.49 | 9.46 | 9.8 | 8.45 | 7.59 | 6.55 | 44.91 | 3.4 | 0.32 |
| Wikn | 0.01 | 0.38 | 0.48 | 1.52 | 2.76 | 3.86 | 5.06 | 5.92 | 6.37 | 6.37 | 51.44 | 13.12 | 2.71 |
| Winer | 0.04 | 0.32 | 0.73 | 1.81 | 2.48 | 2.23 | 3.22 | 4.35 | 5.28 | 6.22 | 55.88 | 14.31 | 3.13 |
| Wnut16 | 0 | 6.2 | 2.66 | 4.87 | 8.12 | 12.56 | 14.62 | 11.08 | 10.64 | 7.53 | 20.54 | 0.9 | 0.28 |
| Wnut17 | 0 | 5.21 | 2.61 | 4.07 | 7.33 | 11.4 | 13.52 | 9.28 | 9.28 | 6.19 | 26.56 | 4.25 | 0.3 |
| Zh | 0.46 | 0 | 7.77 | 0 | 30 | 0 | 18.41 | 0.01 | 11.42 | 0 | 23.87 | 7.8 | 0.26 |

Table 4 records a statistical analysis of the inter-geographic entity distance. It reveals that shorter distances exhibit higher frequencies of occurrence, while longer distances are comparatively less prevalent. Notably, distances of 0 (with a proportion exceeding 50% in *Fned*) or 1 (with a proportion exceeding 25% in *Wiki*) manifest as frequent occurrences, in contrast to distances of 10 (with proportions of 0.28% and 1.73%, respectively). These findings suggest distinct scenarios where the usage of geographic entity varies. In everyday communication, transitions between geographic entities are the norm, whereas more specialized environments like travel or restaurant recommendations involve a frequent succession of geographic references. Conversely, in the specific communication with explicit environmental constraints, geographic entities tend to be more stable and less prone to substitution.

Furthermore, an intriguing observation is the substantial proportion of distances falling into the 31+ category. This indicates a wide range of distance spans, highlighting the extensive variability in the relationship between geographic entities (in contrast to the dimensions of quantity and length). Specifically, the *Wnut17* exhibits an impressive 77.14% proportion of distances surpassing 31, while *Resume*, *Weibo*, *Winer*, and *Wnut16* also demonstrate proportions exceeding 60%. This phenomenon, akin to the famous "negative space" in paintings, reflects the constraints and mutability of incorporating geographic entities within communication or textual contexts.

Table 4: Statistics (%) of inter-geographic entity distances.

| Dataset / Distance | 0 | 1 | 2 | 3 | 4 | 5 | 6 | 7 | 8 | 9 | 10 | 11-20 | 21-30 | 31+ |
|---|---|---|---|---|---|---|---|---|---|---|---|---|---|---|
| Conll2003 | 10.34 | 5.51 | 3.44 | 3.7 | 5.92 | 8.19 | 5.15 | 3.73 | 3.19 | 2.83 | 2.23 | 14.49 | 7.71 | 23.57 |
| De | 15.31 | 6.38 | 4.13 | 3.1 | 2.97 | 2.7 | 2.61 | 2.28 | 2.08 | 1.92 | 1.67 | 11.98 | 8.09 | 34.78 |
| En | 17.9 | 9.35 | 4 | 2.54 | 2.87 | 2.41 | 1.98 | 1.75 | 1.51 | 1.36 | 1.29 | 10.07 | 7.21 | 35.76 |
| Es | 20.83 | 5.72 | 3.55 | 3.89 | 2.68 | 2.23 | 1.81 | 1.53 | 1.39 | 1.33 | 1.23 | 9.31 | 6.81 | 37.69 |
| Fned | 58.73 | 1.95 | 0.79 | 0.61 | 0.41 | 0.4 | 0.22 | 0.39 | 0.37 | 0.28 | 0.28 | 2.4 | 2.46 | 30.71 |
| Fr | 8.13 | 11.08 | 4.69 | 3.99 | 3.18 | 2.82 | 2.23 | 2 | 1.74 | 1.58 | 1.46 | 10.54 | 7.76 | 38.8 |
| It | 12.92 | 7.55 | 5.7 | 4.12 | 3.46 | 2.66 | 2.08 | 1.91 | 1.61 | 1.41 | 1.31 | 10.99 | 6.98 | 37.3 |
| Literature | 0 | 8.37 | 3.44 | 4.03 | 3.36 | 2.89 | 2.94 | 2.47 | 2.66 | 2.15 | 2.26 | 17.47 | 10.82 | 37.14 |
| MSRA | 8.45 | 8.34 | 1.94 | 2.75 | 1.8 | 2.04 | 2.05 | 1.97 | 2.59 | 3 | 2.31 | 16.97 | 10.44 | 35.35 |
| Nl | 17 | 5.79 | 2.94 | 3.03 | 2.65 | 3.31 | 2.78 | 2.19 | 1.98 | 1.77 | 1.57 | 11.89 | 8.37 | 34.73 |
| Pl | 14.73 | 6.39 | 3.66 | 3.32 | 4.03 | 3.15 | 3.2 | 2.33 | 2.14 | 1.95 | 1.81 | 14.18 | 9.47 | 29.64 |
| Pt | 24.82 | 8.17 | 5.25 | 4.59 | 2.67 | 2.3 | 2.01 | 1.71 | 1.57 | 1.31 | 1.16 | 9.39 | 6.86 | 28.19 |
| Ren1998 | 8.61 | 8.02 | 1.85 | 2.7 | 1.7 | 1.98 | 1.99 | 1.87 | 2.66 | 2.93 | 2.11 | 14.79 | 8.07 | 40.72 |
| Ren2014 | 0.03 | 6.83 | 1.6 | 2.03 | 1.71 | 1.85 | 1.79 | 1.79 | 1.83 | 1.67 | 1.72 | 15.66 | 10.63 | 50.86 |
| Restaurant | 0.86 | 1.38 | 2.36 | 4.49 | 7.58 | 8.34 | 8.73 | 6.2 | 6.68 | 5.29 | 3.61 | 21.54 | 8.65 | 14.29 |
| Resume | 0.48 | 6.25 | 0.96 | 0.48 | 2.88 | 0.96 | 1.44 | 0.96 | 0.48 | 1.92 | 1.92 | 10.08 | 10.08 | 61.11 |
| Ru | 17.49 | 5.68 | 3.89 | 3.35 | 3.2 | 2.78 | 2.64 | 1.99 | 1.78 | 1.71 | 1.55 | 11.87 | 8.56 | 33.51 |
| Weibo | 4.11 | 5.48 | 1.37 | 0.68 | 2.74 | 0 | 0 | 0.68 | 0.68 | 1.37 | 0 | 6.84 | 6.15 | 69.9 |
| Wiki | 21.03 | 25.12 | 7.23 | 10.43 | 6.56 | 4.23 | 3.83 | 3.28 | 2.97 | 2.18 | 1.73 | 7.74 | 1.68 | 1.99 |
| Wikn | 1.41 | 3.89 | 4.23 | 3.5 | 2.88 | 2.54 | 2.4 | 2.08 | 1.65 | 1.52 | 1.24 | 10.31 | 6.46 | 55.89 |
| Winer | 1.14 | 3.37 | 4.03 | 3.58 | 2.94 | 2.51 | 2.32 | 1.93 | 1.4 | 1.14 | 1.09 | 6.76 | 3.41 | 64.38 |
| Wnut16 | 8.8 | 3.14 | 2.12 | 1.34 | 1.41 | 0.71 | 0.94 | 0.71 | 1.34 | 0.55 | 1.1 | 8.48 | 5.01 | 64.35 |
| Wnut17 | 5.32 | 4.02 | 2.08 | 1.95 | 1.69 | 0.91 | 1.3 | 0.78 | 0.52 | 0.52 | 0.91 | 2.08 | 0.78 | 77.14 |
| Zh | 7.44 | 7.8 | 4.38 | 2.25 | 1.96 | 1.48 | 2.94 | 2 | 1.61 | 2.04 | 1.58 | 9.48 | 4.9 | 50.14 |

The exploration of laws on the three dimensions offers us an opportunity to delve into the essence of geographic entity. Such ideas have the potential to ignite added insights at the forefront applications within this domain. It conceivably could drive for advancements, push the boundaries and unleash transformative possibilities in the field.

*3.3. Regression & Metric*

Regression is a ubiquitous quantitative analysis way employed to model the intricate relationship between variables by ascertaining the optimal consistency with observed data (Cameron & Windmeijer, 1997). It is commonly used to understand how changes in the independent variables affect the dependent variable and to make predictions or forecasts. Its goal is to estimate the parameters of a mathematical equation (the regression model) that best represents the relationship between the variables. To identify the optimal-fit results, the least squares method is conventionally employed. The sum of the squared vertical deviations between each data point and the fitted one is minimized. This strategy minimizes the collective discrepancy between the observed data and the predicted values. The determination coefficient, denoted as $R^2$ and confined to the interval of 0 to 1, represents a pivotal metric for evaluating the fidelity to the actual data points. Typically, when an $R^2$ exceeds 0.9, it signifies a strong fit between the regression model and the observed data, suggesting a high degree of conformity. An $R^2$ exceeding 0.8 indicates a reasonably acceptable fit, while values below these thresholds may indicate a lack of compliance with the underlying distribution (Linders & Louwerse, 2023). It is important to emphasize that $R^2$ alone (Chicco et al., 2021) is often sufficient to assess the level of agreement between the fitted model and the actual data, and has proven effective in numerous regression analyses (Axtell, 2001; Adamic & Huberman, 2002; Tuzzi et al., 2009; Piantadosi, 2014; Banshal et al., 2022). Assuming $p$ and $q$ are observed and fitted values, respectively, and $N$ is the size of the dataset. Then $R^2$ is calculated by:

$$R^2 = 1 - \frac{\sum_{i=1}^{N}(p_i - q_i)^2}{\sum_{i=1}^{N}(p_i - avg(p_i))^2}$$

While $R^2$ is quite informative (Chicco et al., 2021; Qi & Wang, 2023), it should be acknowledged that real-world data seldom adheres rigorously to distributions. Therefore, to bolster the assessment of acceptability, we also supplement our analysis with additional metrics from diverse perspectives.

The Kullback-Leibler divergence (KL) test is a widely utilized measure to gauge the similarity between two distributions by quantifying the difference between their predicted and actual distributions (Kullback & Leibler, 1951). It ranges from [0, ∞), equating to zero only when the two are identical ($p = q$). A smaller KL signifies a higher degree of consistency (Pérez-Cruz, 2008). Typically, a value less than 0.5 is considered acceptable. KL is calculated by:

$$KL = \sum_{i=1}^{N} p_i \log \frac{p_i}{q_i}$$

Similarly, the Jensen-Shannon divergence (JS) test (Fuglede & Topsoe, 2004), another extensively utilized measure, and ranges within [0, 1]. A lower value indicates a higher degree of concordance. In practice, a value below 0.2 is often deemed appropriate (Endres & Schindelin, 2003). JS is calculated by:

$$JS = \frac{1}{2}\sum_{i=1}^{N} p_i \log \frac{p_i}{p_i + q_i} + \frac{1}{2}\sum_{i=1}^{N} q_i \log \frac{q_i}{p_i + q_i} + \log 2$$

The final metric, the mean absolute percentage error (MAPE), quantifies the discrepancy between fitted and actual values. Its range spans $[0, +\infty)$, with values below 0.5 generally considered reasonable (De-Myttenaere, 2016). MAPE is calculated by:

$$MAPE = \frac{1}{N}\sum_{i=1}^{N} \left| \frac{q_i - p_i}{p_i} \right| \times (100\%)$$

In brief, $R^2$, KL, JS, and MAPE are our metrics, where, $R^2$ occupies a primary position due to its direct reflection of the fit quality, and the subsequent three metrics contribute additional viewpoints that enrich the evaluation scope.

## 4. RESULT

This section introduces the results to the **H1 – H3**, which are presented separately in Section 4.1 – 4.3.

### 4.1. law of geographic entity quantity

Fig.1 exhibits the fitting results of geographic entity quantity frequency-rank $f_{quantity}$-$r$, where the points and lines represent the observed values and fitted ones respectively.

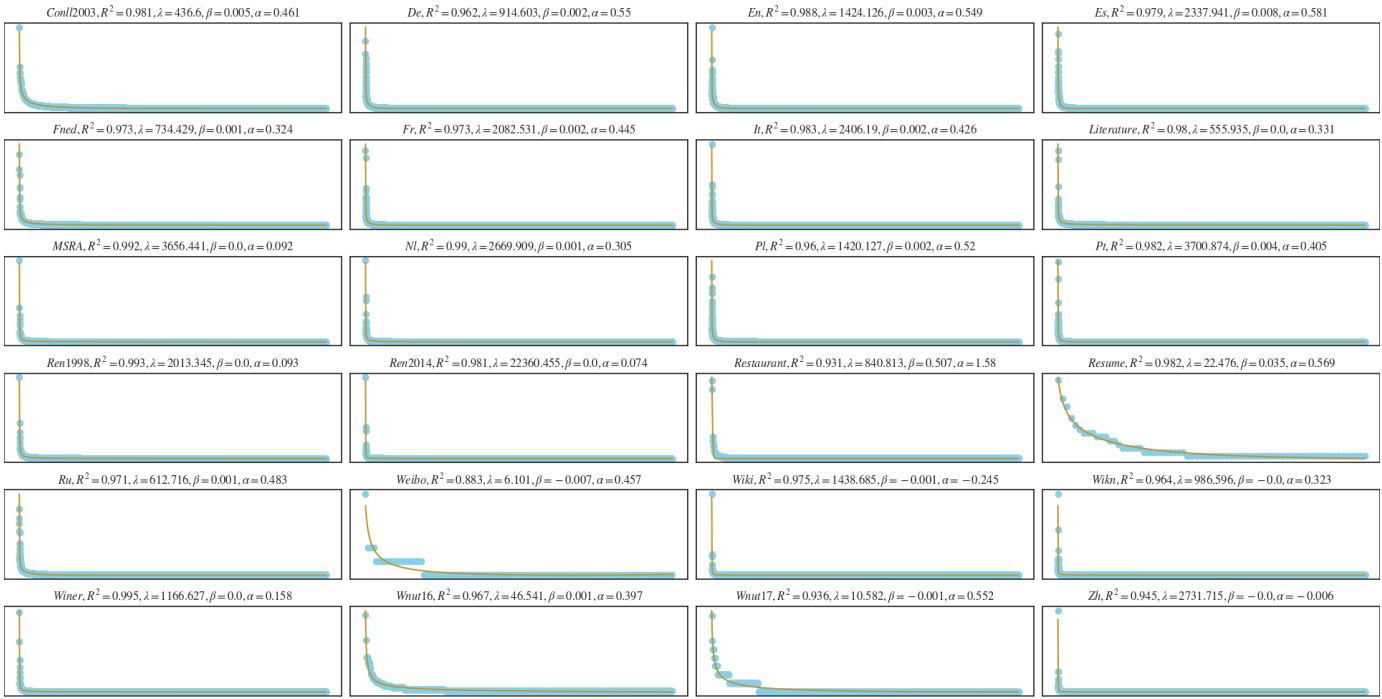

Fig.1: Results of $f_{quantity}$-$r$ on various datasets.

With the exception of the slightly lower $R^2$ of 0.883 observed in *Weibo*, all other datasets exhibit remarkable fitting levels, with $R^2$ exceeding 0.9. In addition, some datasets even demonstrate extraordinary levels of fit, with $R^2$ surpassing 0.99, as exemplified by *Ren1998* and *Winer* etc. Additionally, Table 5 also demonstrates that the overall levels of KL, JS, and MAPE are within acceptable ranges. These findings confirm the suitability of Gamma distribution in capturing $f_{quantity}$-$r$.

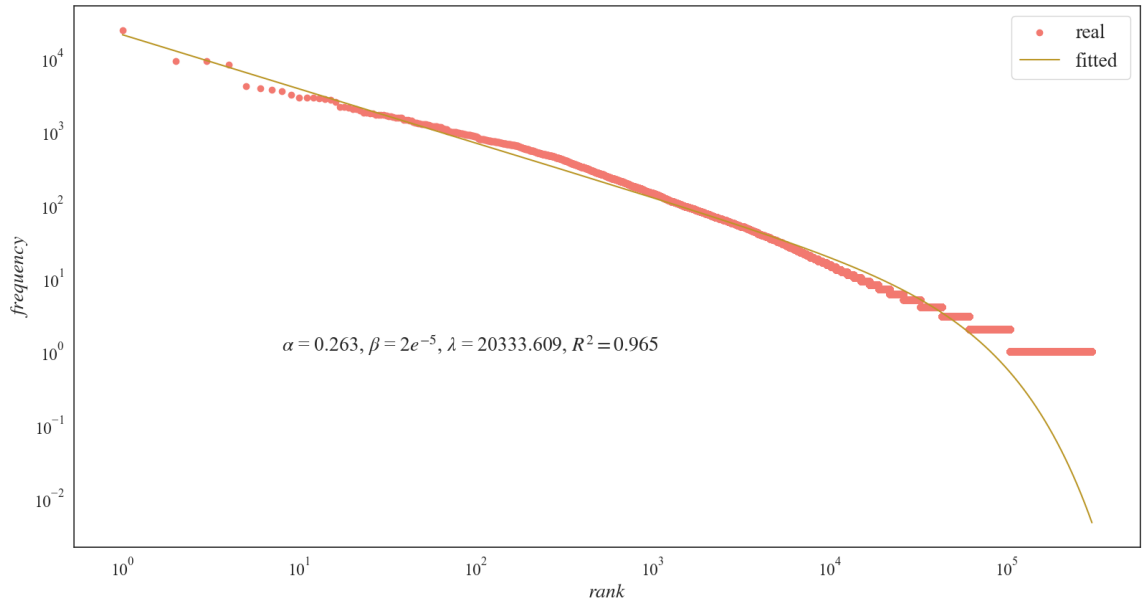

Fig.2: Result of total $f_{quantity}$-$r$.

We should acknowledge that the estimated parameters present significant variability obtained from different datasets. Hence to establish a more robust understanding of this distribution, we employ the total datasets to estimate parameters, as depicted in Fig.2. 0.965 $R^2$ once again confirms our hypothesis. $f_{quantity}$-$r$ can be normalized as Eq.2.

$$f_{quantity}(r) = 0.015 r^{-0.737} e^{2e^{-5}r} \qquad (2)$$

Formally, Eq.2 describes the law of geographic entity quantity. It captures the inverse-like relationship between frequency and rank, as denoted by the -0.737 for $r$ (Note that the exponential function term usually affects the tail without changing the overall trend, hence we can analyze the power function term separately here). This reflects a characteristic of geographic entity, where a few entities dominate frequency and usage, while the majority are less frequently encountered and more scattered. A notable observation emerges from this law, as it deviates from the word frequency distribution ($r$'s power $\approx$ -1) (Piantadosi, 2014) and instead bears resemblance to the urban scaling frequency distribution ($r$'s power $\approx$ -0.7) (Ribeiro et al., 2023). This intriguing distinction can be attributed to the dissimilarities between words and entities within textual contexts. Words predominantly serve as constituent units of text, contributing to the overall expression (Tuzzi et al., 2009). Conversely, entities embody a certain sense and meaning of completeness and integrity, standing as organic wholes that transcend mere word usage (Imai & Gentner, 1997). The dynamics of an entity, encompassing the combination and arrangement of its internal elements, can be seen as a form of simplification akin to the dynamics found within urban scaling that involves the interactions between individuals, although the former exhibits some deviation in scaling invariance. The second part of Eq.2 introduces additional factor that shapes the law. This also reflects the difference between urban and geographic entity, the latter may be more complex.

## 4.2. law of geographic entity length

Laws of length frequency-rank, frequency-length, and length-rank are presented successively.

### 4.2.1. Law of frequency-rank $f_{length}$-$r$

Fig.3 presents the fitting results of $f_{length}$-$r$, with each dataset declaring $R^2$ greater than 0.9, and most exceeding 0.99 indicating almost perfect fit. Also, Table 5 indicates that, on the whole, the values of KL, JS, and MAPE are acceptable. Obviously, $f_{length}$-$r$ follows Gamma distribution.

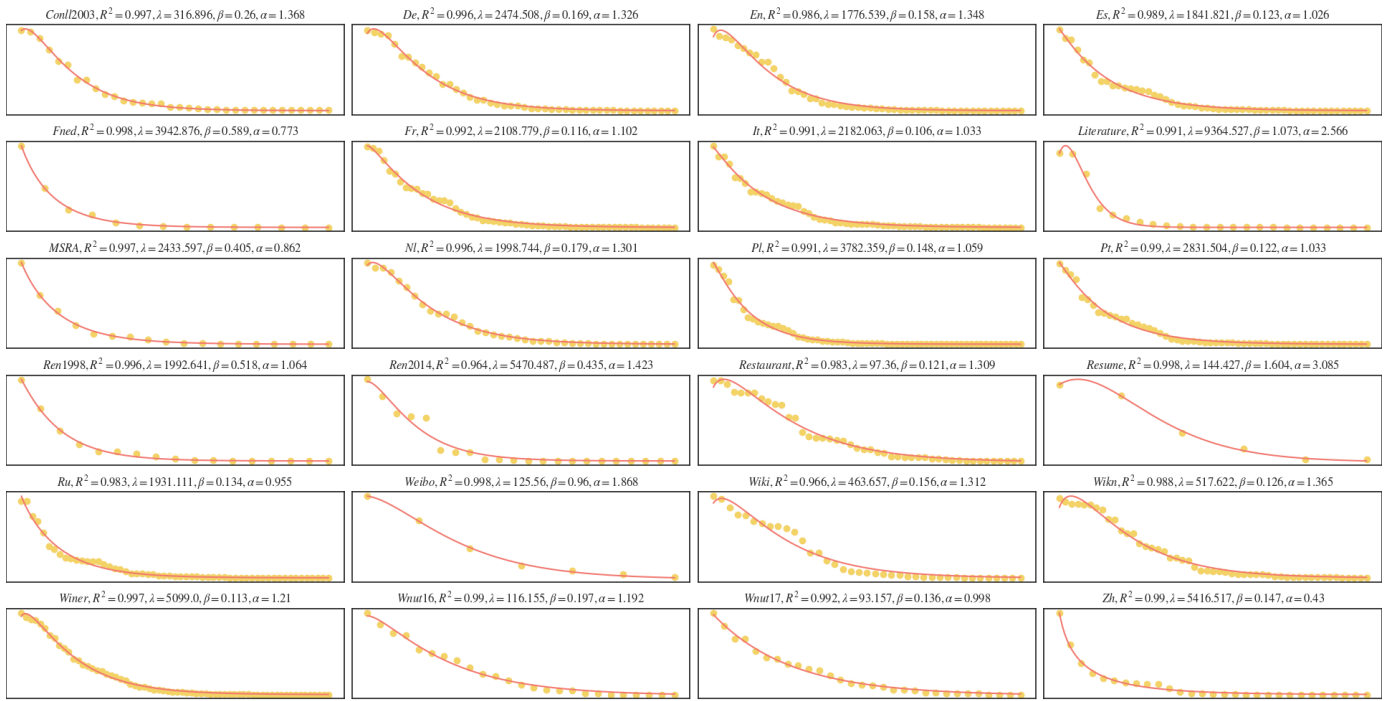

Fig.3: Results of $f_{length}$-$r$ on various datasets.

We use all datasets to fit parameters to alleviate excessive fluctuations in estimated parameters across different datasets, see Fig.4. The $R^2$ of 0.996 also strongly proves the consistency between $f_{length}$-$r$ and Gamma distribution. The law can be normalized as Eq.3

$$f_{length}(r) = 0.018 r^{-0.194} e^{-8e^{-3}r} \qquad (3)$$

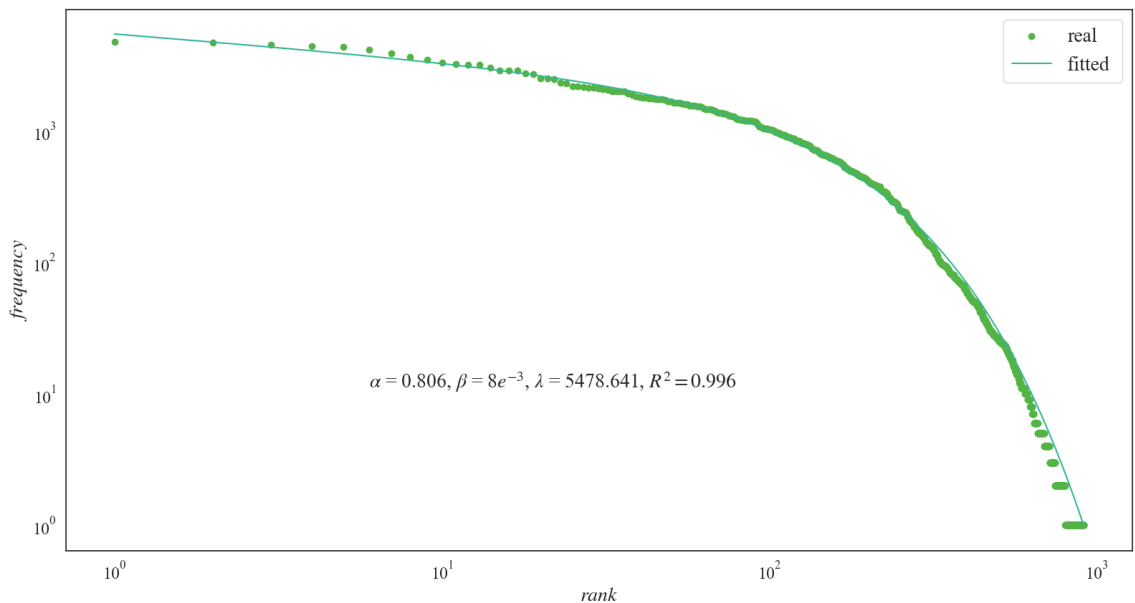

Fig.4: Result of total $f_{length}$-$r$.

This law indicates that as the $r$ increases, its length frequency decreases quasi-exponentially. This suggests that specific lengths of geographic entity are more prevalent and commonly encountered, while a larger number of lengths exhibit rarity and become more dispersed as their rank increases, thereby reflecting the usage dynamics in daily communication. Furthermore, we can re-think Eq.2, an intuitive observation is that difference in parameter *r* between the quantity (-0.737) and length (-0.194) signifies their distinct tendencies. With a higher negative value of *r*, the frequency-rank distribution demonstrates a more rapid decline as the rank increases. -0.737 *r* implies that there is a sharper contrast between the few highly frequent geographic entities and the larger number of less frequently occurring ones. This highlights the phenomenon of concentration and scarcity in the quantity. -0.194 *r* suggests a smoother transition in length from commonly encountered to rarer occurrences, which leans towards the diversity in the length.

### 4.2.2. Law of frequency-length $f_{length}$-$l$

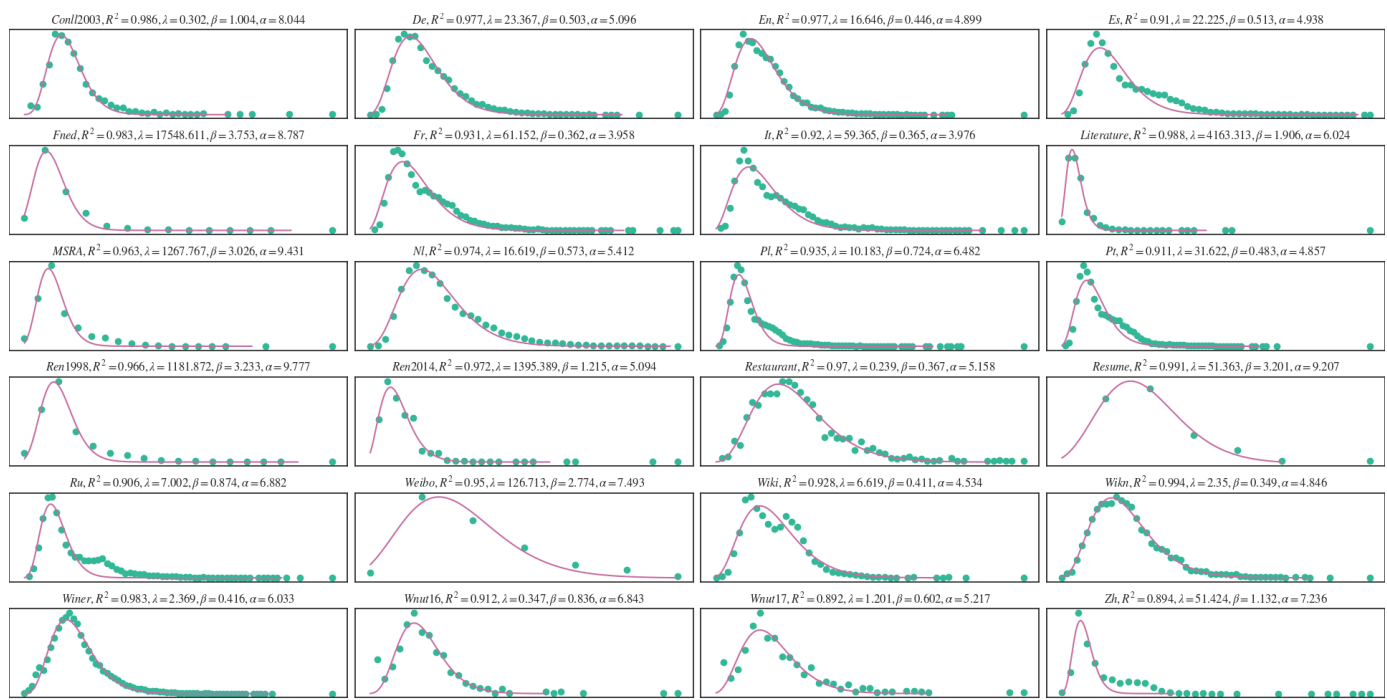

Fig.5: Results of $f_{length}$-$l$ on various datasets.

Fig.5 proves that the observed $f_{length}$-$l$ in each dataset is clearly a Gamma distribution, which is also supported overall by the results of the three metrics in Table 5. Fig.6 integrates all datasets to fit parameters to alleviate the estimated fluctuations, and the $R^2$ of 0.984 is also considerable. The law can be normalized as Eq.4.

$$f_{length}(l) = 0.013 l^{1.792} e^{0.256 l} \quad (4)$$

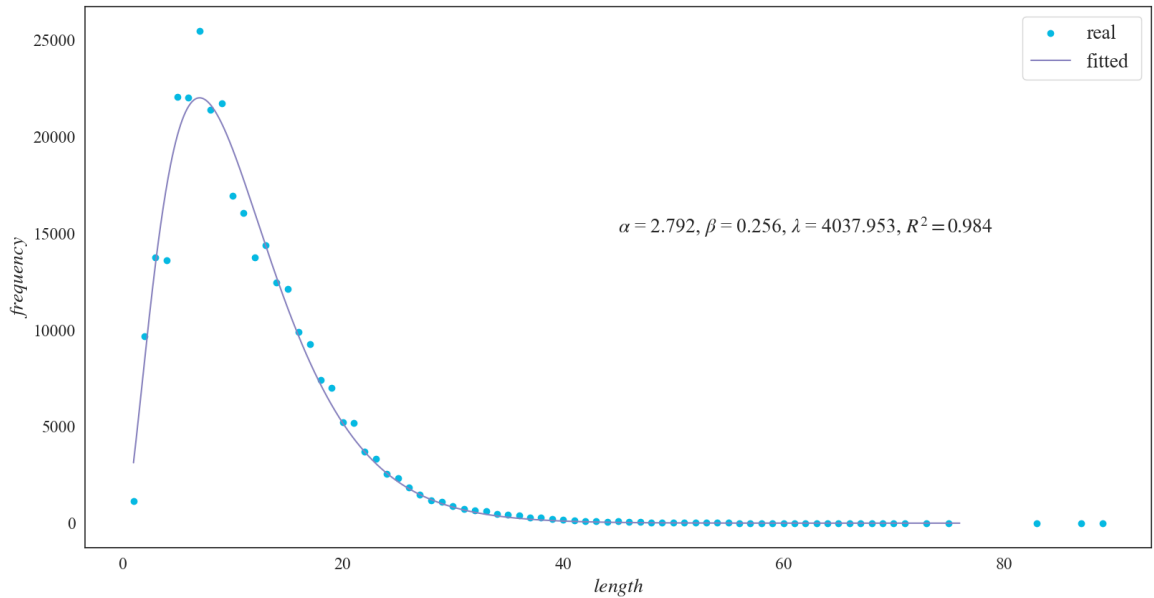

Fig.6: Result of total $f_{length}$-$l$.

The law raises the length variable (*l*) to a positive exponent (1.792). This suggests that, in general, shorter lengths are associated with higher frequencies of occurrence. These parameters imply that there is an initial increase in frequency with shorter lengths, indicating the popularity and common usage of concise geographic entities. However, as the length increases beyond a certain point, the frequency starts to decrease, resulting in a decline in occurrence for longer lengths. This pattern suggests a balancing effect, where extreme lengths are less commonly encountered due to their relative rarity and more specialized nature, highlighting the preference for shorter and more succinct expressions in everyday communication, while acknowledging the existence of longer lengths that are less frequently encountered.

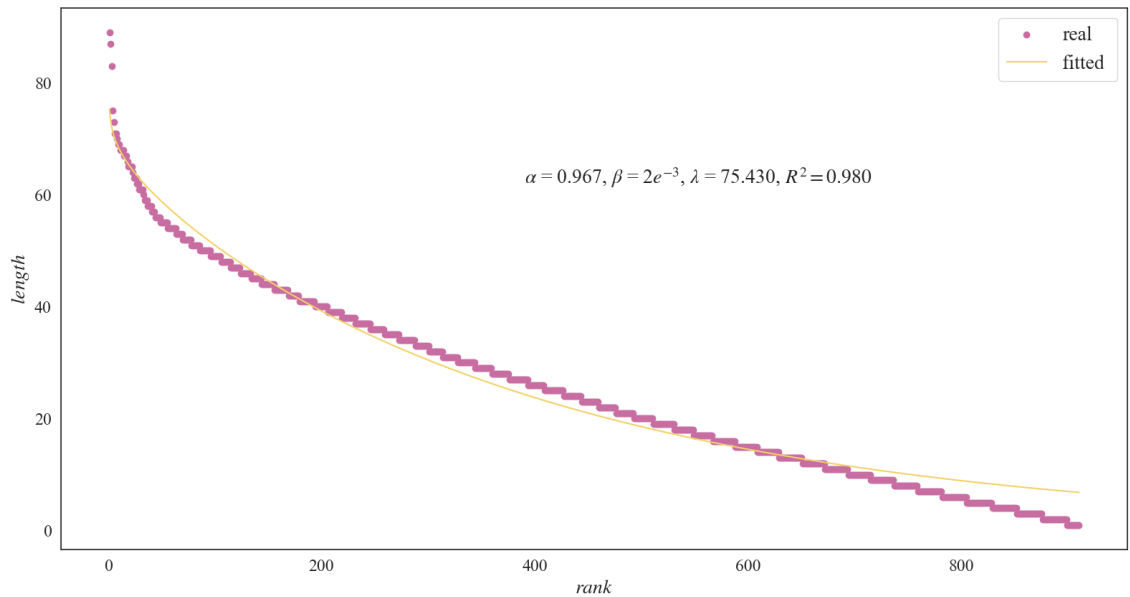

Fig.7: Result of total *l*-*r*.

### 4.2.3. Law of length-rank *l-r*

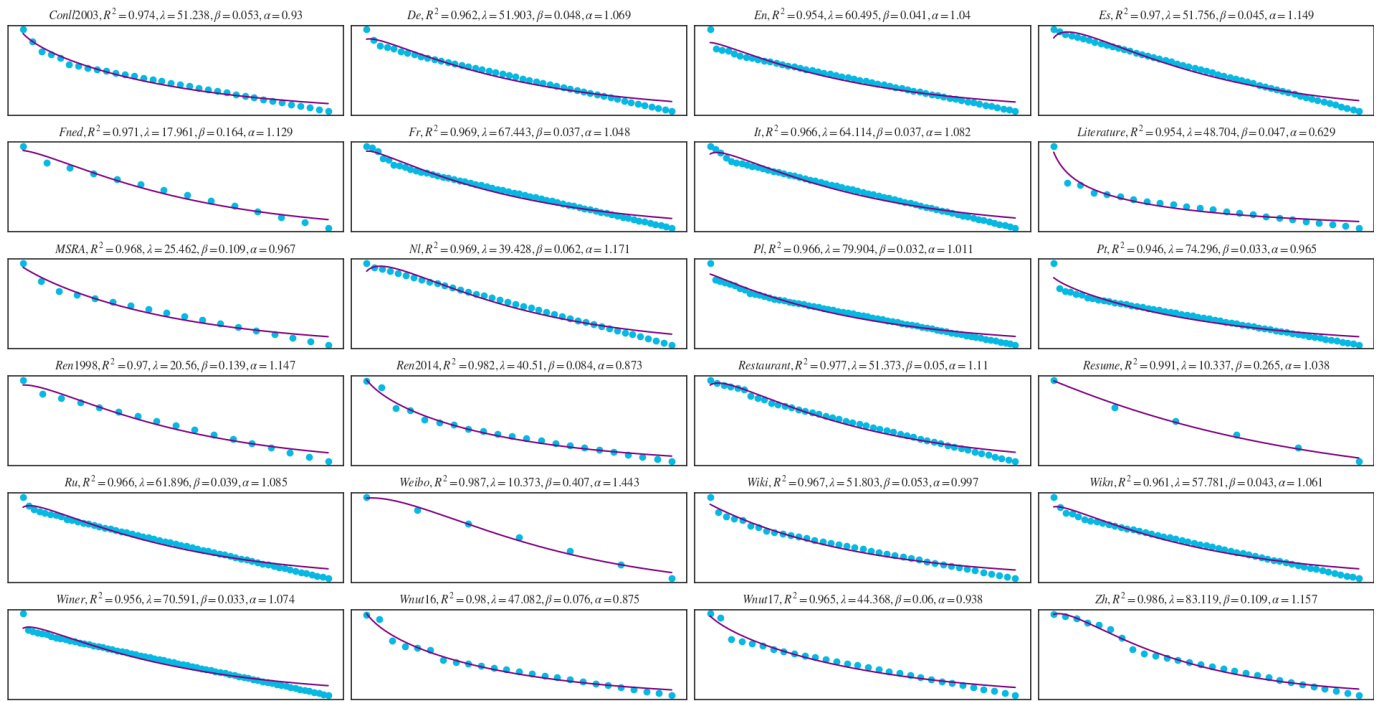

Fig.8: Results of *l-r* on various datasets.

Fig.8 and Table 5 prove that the *l-r* fitting on each dataset is consistent with a Gamma distribution. Fig.7 determines the parameters through all datasets, $R^2 = 0.980$, and the emerging law can be normalized as Eq.5.

$$l(r) = 0.003 r^{-0.033} e^{2e^{-3}r} \qquad (5)$$

The law drives the rank variable (*r*) to a small negative exponent (-0.033). the length undergoes a subtle decrease, and is not drastic, which could exhibit certain persistence across varying levels of rank, indicating that the lengths are analogically resilient. It reveals the adaptability and variability in the expression of geographic entity, accommodating a spectrum of lengths to accommodate various effective communications to a certain extent.

### 4.3. Law of inter-geographic entity distance

We display successively laws of distance frequency-rank, frequency-length, and length-rank.

### 4.3.1. Law of frequency-rank $f_{distance}$-$r$

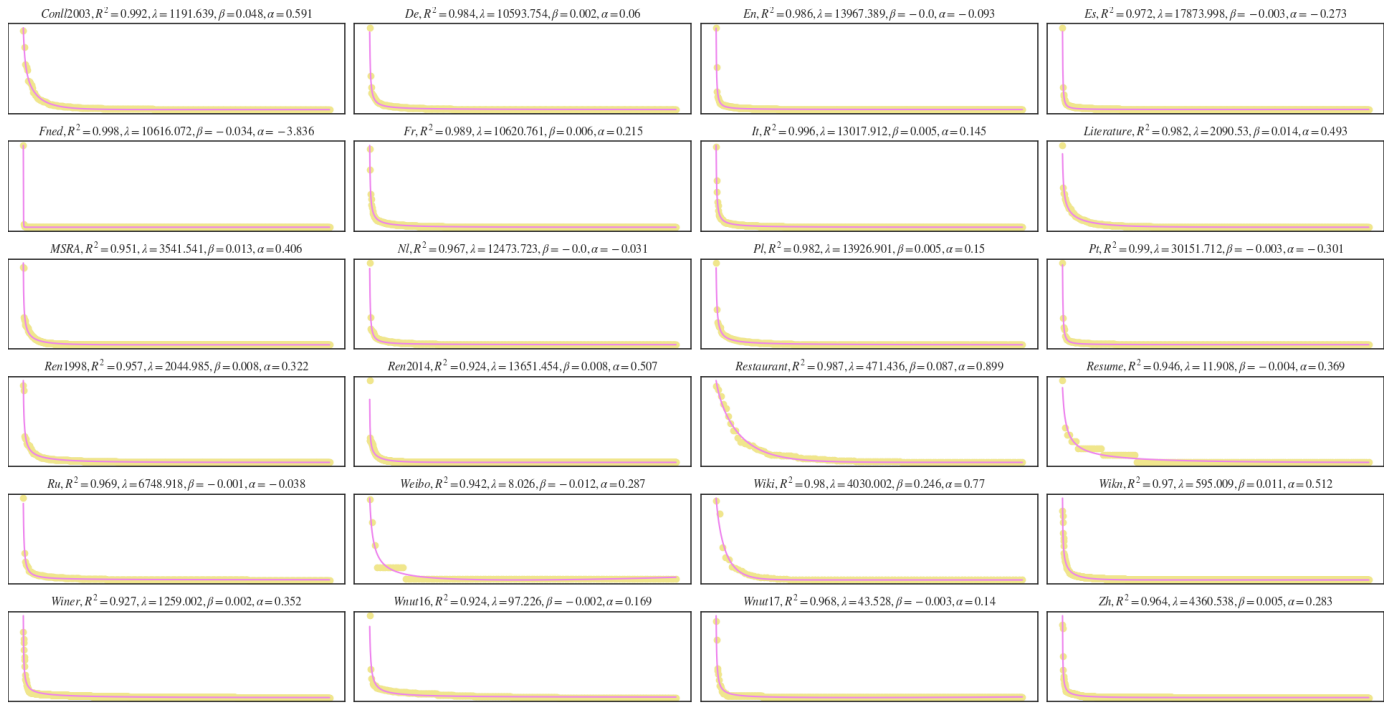

Fig.9: Results of $f_{distance}$-$r$ on various datasets.

Fig.9 assumes that the $f_{distance}$-$r$ fitting on each dataset complies with a Gamma distribution, which is also recognized in Table 5 on the whole. By estimating the parameters on all datasets (see Fig.10, $R^2 = 0.990$), the law can be normalized as Eq.6.

$$f_{distance}(r) = 0.025 r^{-0.566} e^{-e^{-3}r} \qquad (6)$$

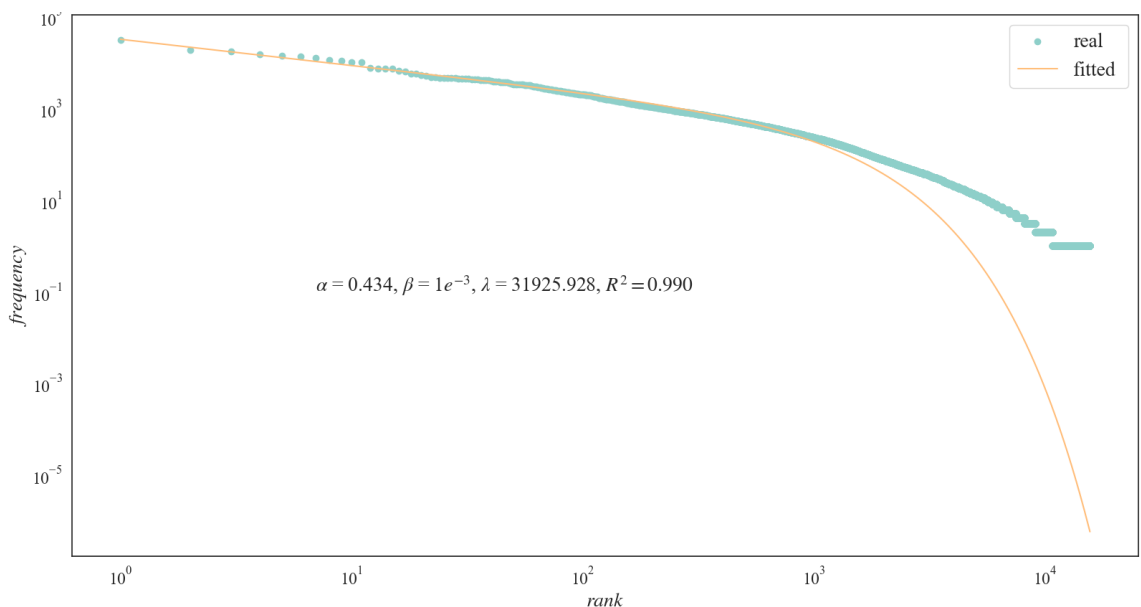

Fig.10: Result of total $f_{distance}$-$r$.

The parameter values signify a phenomenon: the existence of spatial relationships that are frequently utilized in communication. Conversely, a multitude of distances occur infrequently. This indicates a principle of minimum effort, there is a tendency for communicators to opt for distances that require minimal cognitive and memory effort, which aligns with the power limitations of humans and reflects the innate drive to streamline communication by favoring closer and more accessible geographic entity relationships. Compared with rank, the values of $f_{quantity}$-$r$ (Eq.2) and $f_{length}$-$r$ (Eq.3) are -0.737 and -0.194, while that of $f_{distance}$-$r$ is -0.566, which reflects that the decrease in distance with increasing rank is moderate. This suggests that the distances exhibit a relative coherence between popular and common connections, and rarer and more dispersed connections. We can hold a transition from closer and more frequently encountered distances to a wider range of distances, thereby enjoying a greater comfortable sense of control and familiarity over distance for us due to implications for our perception and behavior.

### 4.3.2. Law of frequency-distance $f_{distance}$-$d$

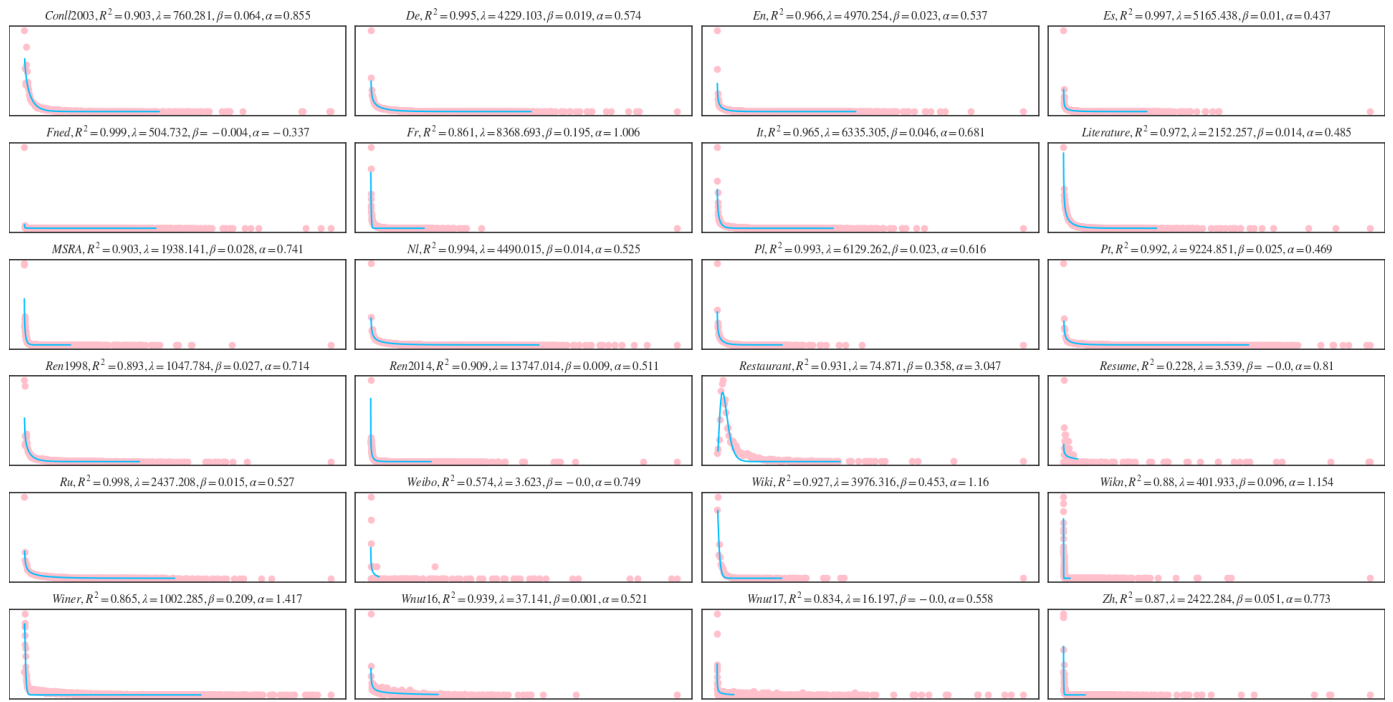

Fig.11: Results of $f_{distance}$-$d$ on various datasets.

With the exception of the *Resume* and *Weibo*, the fitting results on all other datasets demonstrate that the $f_{distance}$-$d$ follows the Gamma distribution see Fig.11. We attribute this deviation to the pronounced fluctuations that occur when the volume is limited. As *Resume* has a total distance frequency sum of only 206, distributed across 136 different distances (206 // 136), and that of *Weibo* is "146 // 121". In comparison with other datasets, such as the first two, the volume of *De* is relatively sufficient – "72,795 // 479", and so is the *Conll2003* – "10,616 // 306". The inevitable deviations in the *Resume* and *Weibo* often lead to more obvious invasions for them. The additional results are shown in Table 5 and are feasible on the whole.

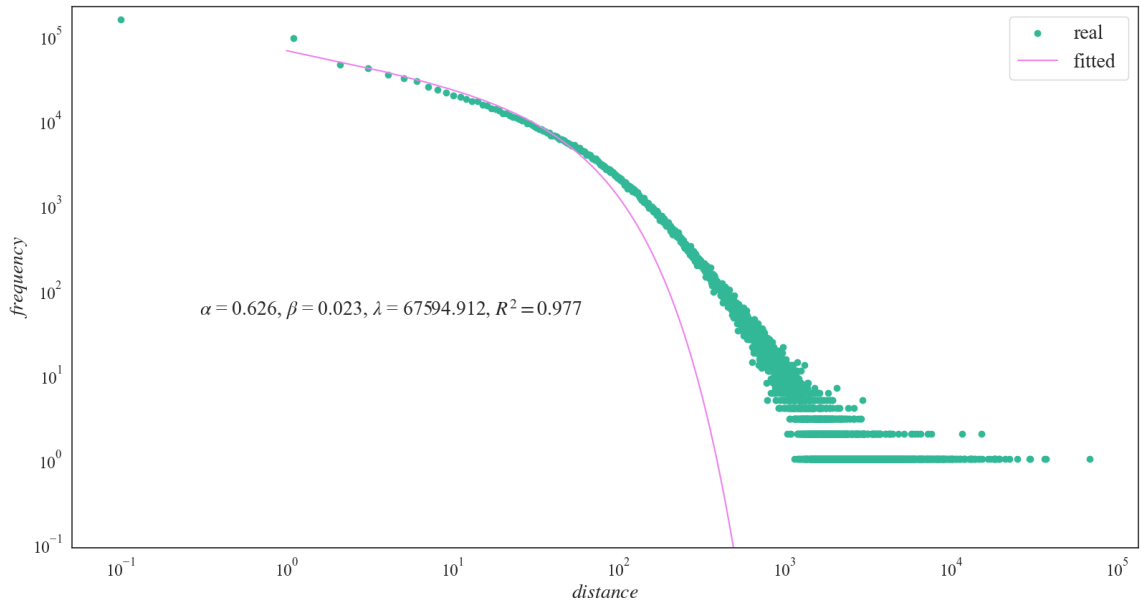

Fig.12: Result of total $f_{distance}$-$d$.

We employ all datasets to estimate the parameters, see Fig.12. The total fitting $R^2 = 0.977$ also confirm this consistency between $f_{distance}$-$d$ and Gamma distribution, and the law can be normalized as Eq.7.

$$f_{distance}(d) = 0.122 d^{-0.847} e^{-0.004d} \qquad (7)$$

The law sheds light on the proximity interplay, suggesting that geographic entities that are closer together exhibits higher popularity and common usage. As the distance increases, indicating greater spatial separation, the frequency of occurrence decreases, resulting in rarer and less commonly used connections. This can be reflected in the term, $d^{-0.847}$, which indicate that the degree of rarity increases as the inter-geographic entity distance grows. The term $e^{-0.004d}$ introduces a decline in the tail of frequency as the distance increases, emphasizing that the rate of decline in frequency becomes more prominent as the degree of spatial separation becomes greater. The preference for closer connections reflects the human tendency to prioritize and rely on nearby spatial relationships. It aligns with concepts of convenience and efficiency, where individuals naturally gravitate towards geographic entity that is easily accessible.

### 4.3.3. Law of distance-rank *d-r*

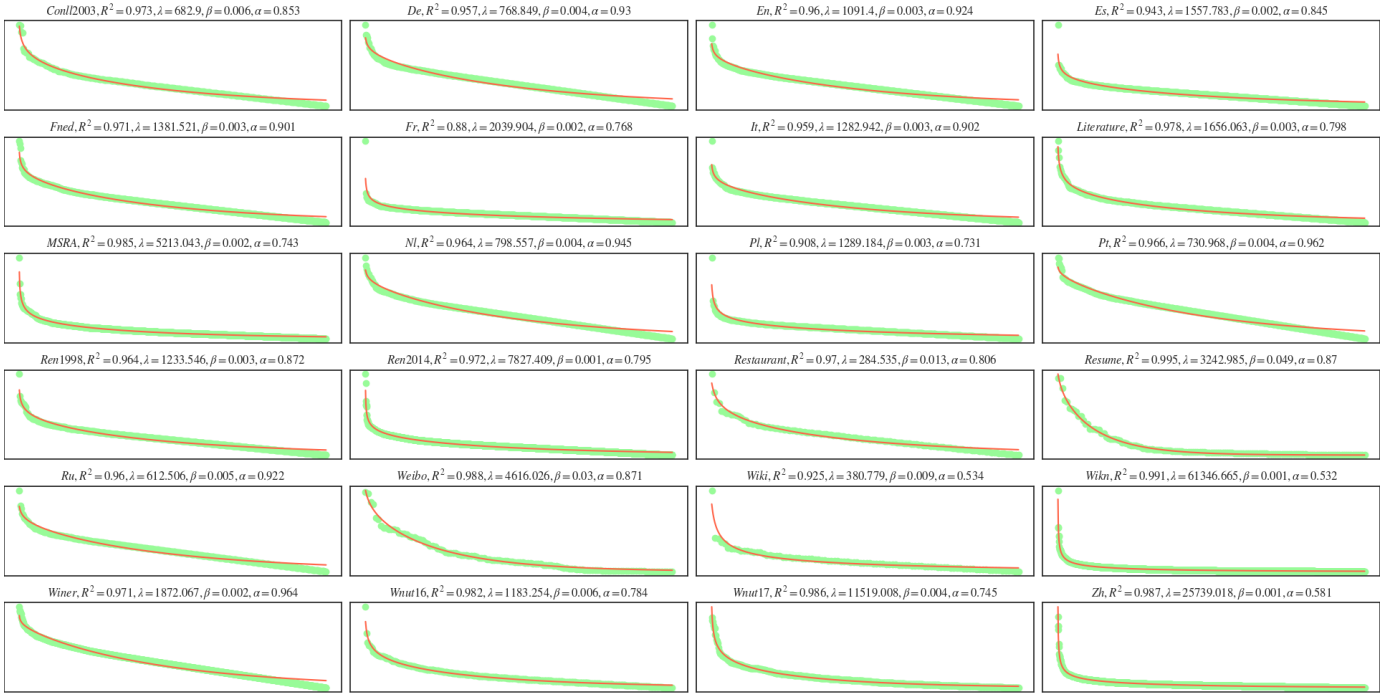

Fig.13: Results of *d-r* on various datasets.

Fig.7 illustrates that the *d-r* fitting on each dataset is consistent with Gamma distribution, with additional acceptability shown in Table 5. Fig.14 determines the parameters through all datasets, $R^2$ is 0.995, and the emerging law can be normalized as Eq.8.

$$d(r) = 0.005 r^{-0.467} e^{-1e^{-4} r} \qquad (8)$$

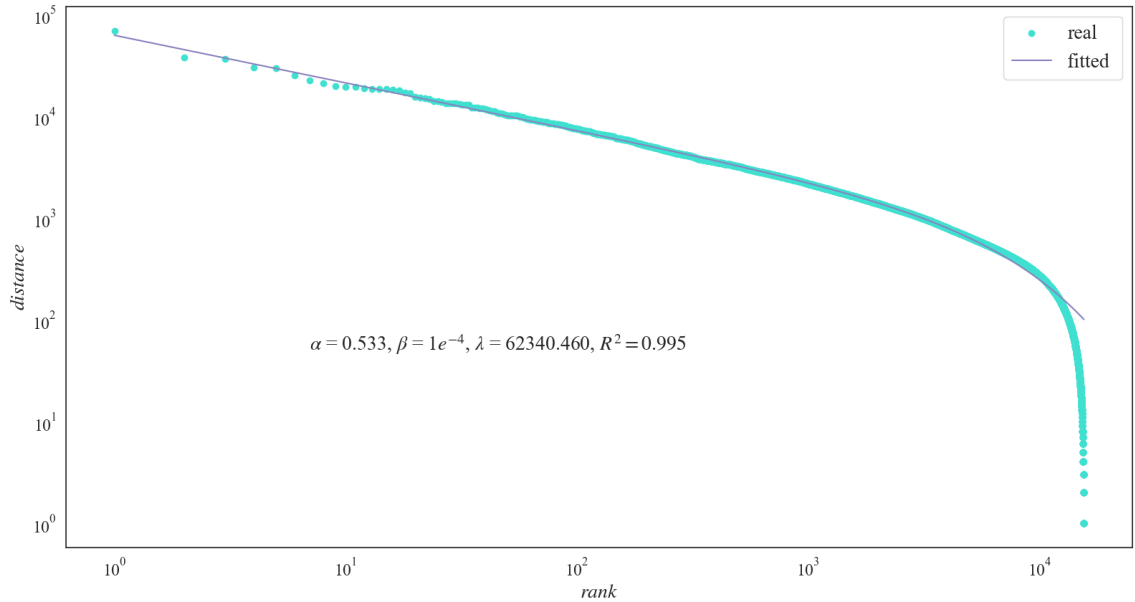

Fig.14: Result of total *d-r*.

The parameter values indicate that the trend of distance variation is not a linear decrease but a rate becomes progressively slower. This phenomenon suggests a saturation effect, where distances reach a point of relative stability as rank increases, the spatial relationships exhibit a level of resilience and stability. It implies that there is a natural limit to how closely connected geographic entities can be, even as their popularity increases. Compared with length-rank, it is evident that the distance experiences a more significant decrease. The stronger concentration of geographic entities in closer distances reflects the preference for space resources efficient utilization, that is in the communication, the proximity may be more crucial.

Table 5: Results of *KL*, *JS* and *MAPE* between different fitted and true distributions.

| | $f_{quantity}$-$r$ | | | $f_{length}$-$r$ | | | $f_{length}$-$l$ | | | $l$-$r$ | | | $f_{distance}$-$r$ | | | $f_{distance}$-$d$ | | | $d$-$r$ | | |
|---|---|---|---|---|---|---|---|---|---|---|---|---|---|---|---|---|---|---|---|---|---|
| | KL | JS | MAPE | KL | JS | MAPE | KL | JS | MAPE | KL | JS | MAPE | KL | JS | MAPE | KL | JS | MAPE | KL | JS | MAPE |
| Conll2003 | 0.027 | 0.009 | 0.269 | 0.003 | 0.001 | 0.105 | 0.070 | 0.016 | 0.381 | 0.006 | 0.002 | 0.125 | 0.117 | 0.027 | 0.600 | 0.458 | 0.069 | 0.832 | 0.012 | 0.005 | 0.255 |
| De | 0.227 | 0.052 | 0.600 | 0.006 | 0.002 | 0.423 | 0.029 | 0.009 | 0.531 | 0.007 | 0.003 | 0.148 | 0.001 | 0.001 | 0.046 | 0.014 | 0.004 | 0.395 | 0.016 | 0.006 | 0.333 |
| En | 0.126 | 0.029 | 0.542 | 0.012 | 0.005 | 0.425 | 0.023 | 0.008 | 0.462 | 0.009 | 0.003 | 0.171 | 0.004 | 0.001 | 0.206 | 0.014 | 0.004 | 0.389 | 0.017 | 0.007 | 0.372 |
| Es | 1.143 | 0.116 | 0.874 | 0.014 | 0.005 | 0.544 | 0.121 | 0.036 | 0.758 | 0.007 | 0.003 | 0.147 | 0.002 | 0.001 | 0.058 | 0.006 | 0.002 | 0.309 | 0.020 | 0.008 | 0.411 |
| Fned | 0.020 | 0.007 | 0.221 | 0.004 | 0.001 | 0.163 | 0.175 | 0.025 | 0.677 | 0.001 | 0.001 | 0.048 | 0.032 | 0.012 | 0.290 | 0.040 | 0.015 | 0.364 | 0.015 | 0.006 | 0.363 |
| Fr | 0.237 | 0.052 | 0.654 | 0.012 | 0.004 | 0.634 | 0.054 | 0.019 | 0.685 | 0.008 | 0.003 | 0.169 | 0.002 | 0.001 | 0.081 | 0.006 | 0.002 | 0.317 | 0.027 | 0.011 | 0.470 |
| It | 0.191 | 0.043 | 0.625 | 0.011 | 0.004 | 0.579 | 0.073 | 0.023 | 0.617 | 0.008 | 0.003 | 0.171 | 0.002 | 0.001 | 0.073 | 0.008 | 0.003 | 0.300 | 0.017 | 0.007 | 0.385 |
| Literature | 0.003 | 0.001 | 0.044 | 0.051 | 0.012 | 0.615 | 0.090 | 0.016 | 0.672 | 0.008 | 0.003 | 0.140 | 0.007 | 0.003 | 0.161 | 0.144 | 0.024 | 0.727 | 0.015 | 0.006 | 0.346 |
| MSRA | 0.010 | 0.004 | 0.248 | 0.006 | 0.002 | 0.296 | 0.313 | 0.050 | 0.684 | 0.003 | 0.001 | 0.074 | 0.030 | 0.010 | 0.387 | 0.920 | 0.071 | 0.894 | 0.012 | 0.005 | 0.388 |
| Nl | 0.028 | 0.009 | 0.306 | 0.004 | 0.001 | 0.271 | 0.040 | 0.012 | 0.531 | 0.005 | 0.002 | 0.117 | 0.003 | 0.001 | 0.121 | 0.021 | 0.006 | 0.467 | 0.015 | 0.006 | 0.328 |
| Pl | 0.561 | 0.098 | 0.754 | 0.007 | 0.002 | 0.103 | 0.236 | 0.048 | 0.773 | 0.009 | 0.004 | 0.185 | 0.012 | 0.004 | 0.368 | 0.005 | 0.002 | 0.325 | 0.025 | 0.010 | 0.405 |
| Pt | 2.158 | 0.145 | 0.887 | 0.013 | 0.005 | 0.502 | 0.109 | 0.032 | 0.745 | 0.011 | 0.004 | 0.198 | 0.001 | 0.001 | 0.067 | 0.004 | 0.001 | 0.276 | 0.014 | 0.006 | 0.314 |
| Ren1998 | 0.012 | 0.005 | 0.282 | 0.010 | 0.003 | 0.199 | 0.316 | 0.050 | 0.674 | 0.002 | 0.001 | 0.058 | 0.014 | 0.005 | 0.174 | 0.820 | 0.110 | 0.865 | 0.016 | 0.007 | 0.361 |
| Ren2014 | 0.015 | 0.006 | 0.199 | 0.045 | 0.017 | 1.744 | 0.020 | 0.007 | 0.343 | 0.003 | 0.001 | 0.075 | 0.542 | 0.058 | 0.864 | 0.539 | 0.060 | 0.855 | 0.018 | 0.007 | 0.553 |
| Restaurant | 4.492 | 0.196 | 0.934 | 0.008 | 0.003 | 0.175 | 0.020 | 0.007 | 0.227 | 0.005 | 0.002 | 0.127 | 0.054 | 0.016 | 0.456 | 0.330 | 0.053 | 0.786 | 0.008 | 0.003 | 0.162 |
| Resume | 0.002 | 0.001 | 0.058 | 0.001 | 0.001 | 0.053 | 0.005 | 0.002 | 0.109 | 0.001 | 0.001 | 0.012 | 0.002 | 0.001 | 0.049 | 0.088 | 0.032 | 0.334 | 0.018 | 0.006 | 0.320 |
| Ru | 0.031 | 0.010 | 0.277 | 0.012 | 0.004 | 0.169 | 0.474 | 0.081 | 0.746 | 0.008 | 0.003 | 0.166 | 0.002 | 0.001 | 0.115 | 0.006 | 0.002 | 0.306 | 0.015 | 0.006 | 0.294 |
| Weibo | 0.001 | 0.001 | 0.030 | 0.001 | 0.001 | 0.063 | 0.001 | 0.001 | 0.063 | 0.001 | 0.001 | 0.004 | 0.001 | 0.001 | 0.001 | 0.031 | 0.011 | 0.129 | 0.018 | 0.007 | 0.874 |
| Wiki | 0.027 | 0.010 | 0.223 | 0.032 | 0.012 | 0.786 | 0.030 | 0.011 | 0.399 | 0.006 | 0.002 | 0.132 | 0.095 | 0.020 | 0.710 | 0.080 | 0.015 | 0.653 | 0.017 | 0.007 | 0.220 |
| Wikn | 0.009 | 0.003 | 0.091 | 0.010 | 0.004 | 0.422 | 0.008 | 0.002 | 0.179 | 0.008 | 0.003 | 0.160 | 0.117 | 0.033 | 0.532 | 0.116 | 0.032 | 0.536 | 0.009 | 0.004 | 0.526 |
| Winer | 0.004 | 0.001 | 0.084 | 0.009 | 0.004 | 1.124 | 0.030 | 0.009 | 0.457 | 0.010 | 0.004 | 0.190 | 0.030 | 0.011 | 0.157 | 0.045 | 0.017 | 0.692 | 4.396 | 3.772 | 0.999 |
| Wnut16 | 0.005 | 0.002 | 0.082 | 0.006 | 0.002 | 0.155 | 0.172 | 0.031 | 0.209 | 0.003 | 0.001 | 0.086 | 0.029 | 0.011 | 0.227 | 0.095 | 0.034 | 0.444 | 0.009 | 0.003 | 0.227 |
| Wnut17 | 0.002 | 0.001 | 0.054 | 0.004 | 0.001 | 0.114 | 0.109 | 0.028 | 0.278 | 0.006 | 0.002 | 0.120 | 0.007 | 0.002 | 0.102 | 0.067 | 0.023 | 0.262 | 0.010 | 0.004 | 0.525 |
| Zh | 0.083 | 0.031 | 0.432 | 0.036 | 0.015 | 2.701 | 1.173 | 0.125 | 0.733 | 0.003 | 0.001 | 0.097 | 0.018 | 0.006 | 0.264 | 0.133 | 0.033 | 0.626 | 0.012 | 0.005 | 0.472 |

## 5. DISCCUSSION

### 5.1. Upper-boundary estimation

De et al., (2021) argue that dynamical systems reflected in the distribution of objects often exhibit temporary under-sampling (limited statistical data is inevitable), leading to instability, which manifested as deviations in object scale, allows us to estimate potential upper-cutoff $o_M$ in the system. Building upon this insight, we have an opportunity to explore the boundaries of geographic entity.

$\lambda$ is generally regarded as a free parameter, and the empirical deviation $D_e$ satisfies:

$$D_e = N(o_m / o_M)^{1/1-\alpha} e^{\beta(o_m - o_M)}$$

Where, $N$ is the number of objects, $o_m$ is the lower-cutoff. The initial value used for iterating $o_M$ is given by the largest observed object $oo_M$ in the system.

Then follow:

$$o_M = o_m (1 + D_e / N)^{1/1-\alpha}$$

Since α ≠ 0, allows $D_e$ ≠ 1, thereby implying that the $o_M$ does not exhibit divergence. By iterating the aforementioned equations, we can estimate it. Specific proof can be found in this research (De et al., 2021). The estimations are presented in Table 6, where, the column $D_e$ records its average value, $oo_M$ is the observed upper-cutoff.

Table 6: Results of upper-boundary estimation.

| Parameter / Distribution | $f_{quantity}$-r | $f_{length}$-r | f-l | l-r | $f_{distance}$-r | f-d | d-r |
|---|---|---|---|---|---|---|---|
| $D_e$ | 20181277 | 965 | 0 | 900 | 61426 | 3720 | inf |
| $o_M$ | 47972 | 9316 | 25467 | 118 | 37571 | 184838 | 74867 |
| $oo_M$ | 23368 | 4754 | 25467 | 89 | 30987 | 154630 | 68756 |

It is worth noting that a low value of $D_e$ is evident in the *f-l* distribution, indicating relative stability in this system. This affords us to estimate $o_M$ directly from solving the maximum value of the *f-l* law, conditioned on that *l* is a positive integer $N^+$ and in conjunction with observed values.

$$o_M = \max(oo_M, \max(f\text{-}l)), l \in N^+$$

We can gain the following viable speculative conclusions:

Quantity of geographic entity usage: The estimation that the maximum frequency of geographic entity usage is 47972, driven by the pursuit of communication efficiency. However, the existing dataset's observed peak of 23368, which is approximately half of the estimated $o_M$, suggests the presence of additional factors influencing geographic entity frequencies. One possibility is that geographic entities with frequencies surpassing 23368 may exist in other domains, such as spoken language texts. In these domains, the finer division or granularity of geographic entity might not be as prominent, resulting in multiple geographic references converging onto the same entity. Consequently, the usage frequency of such entities could significantly surpass the observed peak $oo_M$. Another intriguing conjecture is that there may be unexplored regions or uncharted territories yet to be fully integrated into communication systems. In this context, it is conceivable that future human interactions might involve more diverse and crazy geographic entities from the universe, expanding the frontiers of communication and knowledge exchange.

Length preference: The understanding of geographic entity length reveals fascinating patterns that reflect human cognitive preferences and the underlying cognitive load. The consistent (25467) $oo_M$ and $o_M$ in length frequency-rank distribution indicate the existence of a robust preference for certain lengths of geographic entity in current communication practices. The frequency of 9316 can signify the extent of this preference. Conversely, the observed peak $oo_M$ of 4754 implies the existence of a substantial reservoir of unknown geographic entities with different lengths yet to be explored and integrated into communication systems. The presence of a maximum length of 119, slightly

25higher than the observed value of 89, may indicate the presence of specialized geographic entities that have not been adequately captured or observed, including the existence of uncharted territories or unconsidered domains that contain 20 longer geographic entities. This insight raises another promising idea, the speaker and listener's maximum willingness threshold for the length stands at 119, beyond which a sense of fatigue sets in. This stems from the realization that there is no imperative to articulate comprehensive geographic details exceeding this limit in one instance. Even when dealing with intertwined or coupled information with multiple geographic references, such complex data can be effectively broken down into a sequence of shorter entities for expression. An illustrative extreme is that we may posit that geographic information with a length of 120 could conceivably be disassembled into 60 entities, each with a length of 2, and subsequently articulated. Naturally, the specific form of decomposition is contingent on the laws of quantity (see Eq.2). In view of the above, an intuitive engineering practice emerges for the extraction of geographic entities in uncharted domains where an optimization condition is that the gains achieved through a sequence of spans, conforming to a gamma distribution, remain at or below the critical threshold of 119.

Distance in communication strategy: The examination of distance distributions in geographic entity usage provides insights into the underlying communication strategies employed by humans. The close proximity between the estimated upper-cutoff and the observed peak from the three laws implies a saturation point in human acceptance of different distances when expressing geographic entity. This indicates that current communication practices have almost accounted for various distance preferences, filling the existing gaps and unknowns, particularly in scenarios where geographic entity is constrained or predefined. Furthermore, we can speculate that humans could develop a communication threshold when it comes to distance. The estimated maximum distance of 74867 suggests that once a geographic entity is introduced in a conversation, human cognition "compels" the next geographic entity instance to be referenced within 74867 words at the latest. This cognitive inclination ensures the continuity of conversation, comprehension of context, and overall efficiency in information exchange. Understanding the dynamics of distance provides valuable insights for optimizing geographic information engineering. By accounting for distance preferences, one improvement is that we can discover hidden geographic information and have a greater chance of extracting it.

These insights are profound and have the potential to deepen our contemplation of geographic information science.

**5.2. Theoretical analysis**

Geographic entity quantity utilization, length consideration and distance selection, embody the intricacies of human cognition and ratiocination, intrinsically entwined with life's processes. Within this section, distance selection serves as a case in point to expound upon the theoretical underpinnings, whereby its distribution assumes a pivotal role in elucidating the salient facets delineating the "lifecycle" of unbroken sequences of geographic entities. Thus, we analyze the main factor behind the emergence of Gamma distribution's contour to encapsulate the dynamics intrinsic to the uninterrupted progression of geographic entity.

Assuming the process of embedding geographic entities within a text. After each geographic entity is determined, the speaker probabilistically places the next geographic entity with a probability of $p$, while the probability of not placing it is $1-p$. Each decision regarding the placement of geographic entity follows a Bernoulli trial characterized by $f(x) = p^x(1-p)^{1-x}$. Therefore, the collective outcome of $n$ consecutive decisions satisfies:

$$\lim_{n \to \infty} \left[ \binom{n}{x} p^x (1-p)^{n-x} \right]$$

Due to the expected value of the terms within the square brackets being $E(X) = np = \beta$, the consecutive placement of $n$ items can be represented by:

$$\lim_{n \to \infty} \binom{n}{x} \frac{\beta^x}{n} (1 - \frac{\beta}{n})^{n-x} = \frac{\beta^x}{x!} e^{-\beta}$$

Then the distance between consecutive geographic entities is the number $t$ ($t = 0, 1, 2, \ldots$) of words between them required to achieve first success, following

$$1 - e^{-\beta t}$$

The trial with a success in this process follows an exponential distribution with parameter $\beta$. Obviously, the distribution of $\alpha$ successes satisfies:

$$\frac{\beta^{\alpha+1}}{\Gamma(\alpha+1)} t^\alpha e^{-\beta t}$$

Which follows the probability density function of *Gamma*($\alpha + 1, \beta$).

End of analysis.

**5.3. Comparison with Gaussian distribution and Zipf's law**

We shall incorporate Gaussian distribution and Zipf's law into comparative evaluations. Where, Gaussian distribution is commonly used for quantifying phenomena in natural and behavioral sciences (Altman & Bland 1995).

Due to the fact that geographic entities are general objects, it is worth measuring whether they can be characterized by Gaussian distributions and to what extent. Considering that Zipf's law manifests prominently in the frequency distribution of words in natural language (Linders & Louwerse, 2023). Since entities are composed of words, it is essential to evaluate whether geographic entity aligns with this law.

Gaussian distribution and Zipf's distribution model geographic entities as Eq.9-10, where δ, C, σ and μ can be estimated using regression.

$$f(x) = \frac{1}{\sqrt{2\pi}\sigma} e^{\frac{(x-\mu)^2}{2\sigma^2}} \qquad (9)$$

$$f(x) = \frac{C}{x^{\delta}} \qquad (10)$$

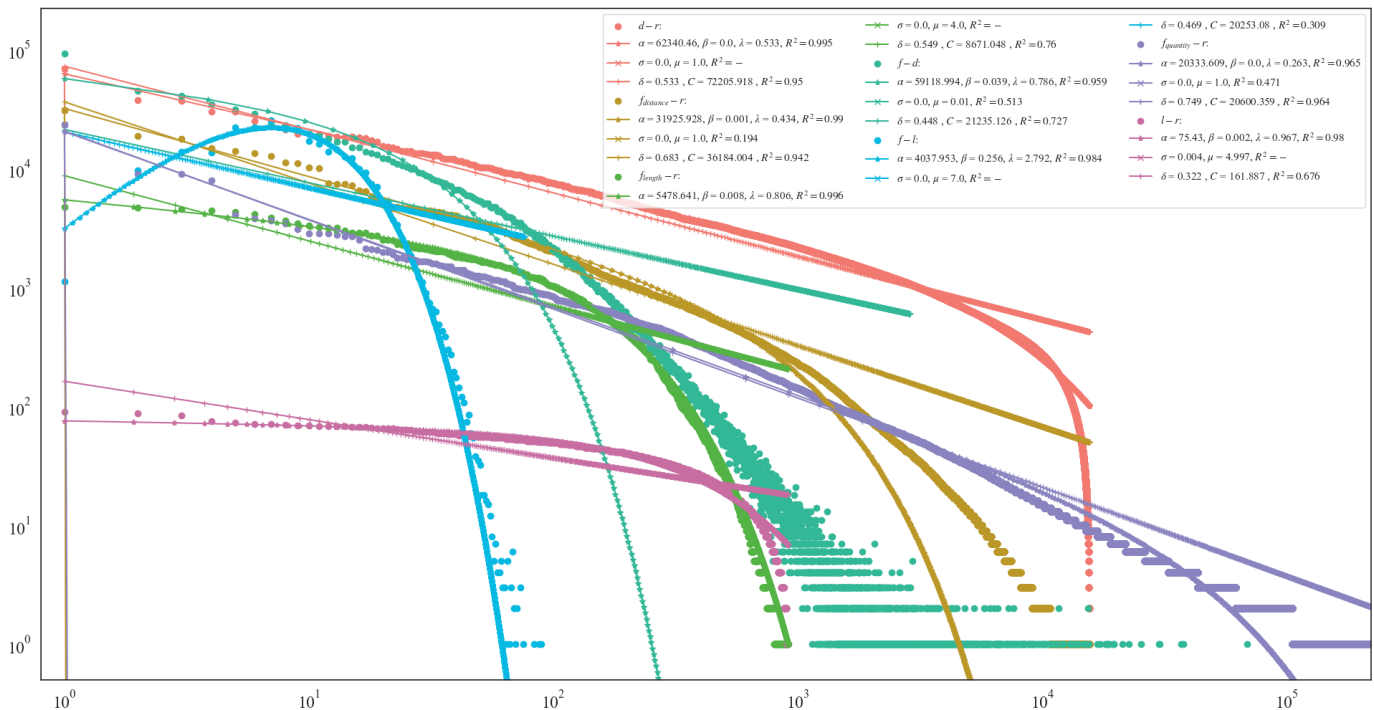

Fig.15: "●", "★", "×" and "|" represent the value of real, Gamma distribution fitted, Gaussian distribution fitted, and Zipf distribution fitted, respectively.

The respective $R^2$ of 0.194, 0.513, and 0.471 for $f_{distance}$-r, f-d and $f_{quantity}$-r unequivocally indicate a substantial deviation of geographic entity from the Gaussian distribution. This departure is particularly evident in d-r, $f_{length}$-r, f-l and l-r, where the significant disparities between the actual values and those predicted by Gaussian fitting result in the vanishing of $R^2$, thereby vehemently rejecting the compatibility of Gaussian distribution with geographic entity. This nonconformity serves as a testament to the fact that the utilization of geographic entity transcends randomness and indeed encapsulates the intricate interplay of human cognition. This phenomenon, related to the life, lies outside the purview of the Gaussian distribution and, instead, introduces a new member to the family of Gamma distribution. For effective communication to take place, it becomes imperative that certain geographic entity exhibits a concentration towards the upper extremities in terms of quantity, length, and distance, deliberately deviating from the expectations set by Gaussian distribution. This deliberate pattern defies the common notion of massive data aligning around the average. Conversely, there are geographical entities that exhibit dispersion, exhibiting a long tail phenomenon. We can gain insights from the following perspectives as well: As the geographic entities become increasingly distant, speakers may experience heightened fatigue when attempting to recall or refer to those entities, given the limitations of human cognition. Conversely, listeners may also encounter fatigue stemming from the waiting process. Likewise, the lengthier or more the geographic entity, the more likely speakers are to experience fatigue when articulating it. Thus, their frequencies and their own values reject to become a normal bell-shaped trend towards Gaussian distribution, but instead show a decreasing trend overall.

Zipf's law demonstrates commendable performance in scenarios involving $d$-$r$, $f_{distance}$-$r$ and $f$-$r$. Nevertheless, its goodness of fit remains inferior to that of Gamma distribution. Conversely, in the remaining distribution scenarios, Zipf's law proves inadequate. An evident observation arises in the $f$-$l$ scenario, as illustrated by the blue curve in the Fig.15. The fitted Zipf distribution intersects the arc-shaped curve representing the true values, rather than conforming to it. We believe that Zipf's law is unquestionable at the level of characterizing words, while geographic entity encompasses a broader spectrum of meanings. Moreover, it is crucial to acknowledge the overlap between meanings, which, in turn, can give rise to the repeated use of certain words. Take, for instance, the geographic reference to Beijing, which may be expressed in multiple ways, such as "in the northern part of the North China Plain, adjacent to Tianjin in the east" or "116 ° 20 ′ East longitude, 39 ° 56 ′ North latitude". This phenomenon hinders the distribution of entities from an inverse pattern akin to words, where the word with first rank is $n$ times more frequently that of the $n$-th word. Consequently, geographic entity's manifestation within the framework of Zipf's law is inherently biased, thus deviating from conformity.

In summary, the conformity of geographic entity to Gamma distribution is not a random occurrence or accidental phenomenon. The alignment is a meaningful and significant finding.

# 6. IMPLICATION

Through our meticulous exploration, we unveil deterministic laws governing geographic information. These revelations not only enrich our comprehension of geographic information, propelling scientific advancement, but also bear the following potential to serve the extraction of geographic information (GIE) from textual sources.

Firstly, GIE often involves utilizing deep learning techniques, where sequence annotation plays a base (Hu et al., 2022c; Kersten et al., 2022). However, ensuring precise start and end indices for the extracted spans poses challenges in model predictions. In tackling this issue, the length law and distance law come to the fore. For instance, these laws can furnish optimization conditions within the model's loss function, guiding the model in inferring the start index of the subsequent prediction by considering the distances between current output predictions, aided by Eq.8. Together with Eq.5, the model can determine the end index of the prediction. This can enhance accuracy of GIE. Moreover, in certain scenarios, imbalanced categories of geographic entities can exert a considerable influence on the model's performance (Kang et al., 2022). To address this challenge, a viable solution emerges through leveraging Eq.2, which guides the generative model to produce geographical entities with fewer categories. As Eq.2 delineates the distribution of common and rare geographic entities, GPT2 can effectively generate the less frequent ones, effectively expanding their scale. This approach ensures category balance and fosters GIE. Furthermore, the emergence of contrastive learning presents a promising avenue for unsupervised GIE, where the construction of sample-pairs playing a pivotal role (Oh et al., 2023). Leveraging Eq.6-8, we can sample multiple negative instances for each positive sample, resulting in a more reasonable sample-pairs construction. This idea fortifies the capabilities of contrastive learning in GIE, unlocking insights from unannotated data. Additionally, the issue of out-of-vocabulary (OOV) poses a common challenge for models (Keller et al., 2009). One promising approach to address this lies in leveraging Eq.2 to identify potential omissions in GIE. Should such gaps be detected, we can then employ the combination of Eq.4 and Eq.7 to predict the plausible positions of concealed geographic entities, thereby mitigating the OOV problem. This concept bears resemblance to star prediction in astronomy, where hidden celestial bodies can be inferred based on observed patterns, for example, Neptune was calculated by Titius-Bode law (Nieto, 2014). Also, to devise a more robust approach for nested GIE (Finkel & Manning, 2009), there is a motivation: incomplete extraction results in a divergence from the length frequency depicted by Eq.4. To illustrate, consider the nested geographic entity "intersection of Yangtze River Road and Yellow River Road." If this entity is extracted as three separate entities, namely "Yangtze

30River Road," "Yellow River Road," and "intersection," a noticeable deviation from the length laws would ensue. Hence, we can leverage graph-based methods to capitalize on the length distribution characteristics and aid the model in achieving comprehensive extraction. Another concern is that text sources often contain incomplete or unclear geographical references (Qiu et al., 2019). By utilizing the distribution curve in our length laws Eq.3-5, researchers can use statistical inference, contextual, and probabilistic inference to reconstruct missing or uncertain geographic details. Finally, the integration of different geographic datasets typically requires addressing differences and coordinating different representations of geographic entities (Buccella et al., 2009). In this regard, our findings can be embedded on the basis of graph neural networks and probabilistic record links to align the distribution of various geographic datasets, thereby improving the coordination of location references from multiple sources.

All in all, the revelations of these laws usher in a level of expanded boundaries in knowledge, and propel the precision, scalability, and efficacy of GIE to promising heights.

## 7. CONCLUSION

Albert Einstein once remarked, " The essence of truth is the freedom to think for oneself." Isaac Newton, too, reflected upon the matter, asserting that " Nature is pleased with simplicity. And nature is no dummy." These declarations imply that laws serve as immutable markers in our quest to comprehend the world around us.

In this paper, through a regression analysis conducted on 24 diverse datasets encompassing different languages and types, we have elucidated 7 laws governing geographic information, all pointing towards a common direction: Gamma distribution with distinct parameters. Moreover, considering the limitations of human cognition, we estimate the upper-cutoff of geographic information. Apart from the frequency-length relationship, the other distributions reveal that some communication scenarios are yet explored by humanity. However, we should acknowledge that the estimated values may have some imperfections. Additionally, changes or expansions in the datasets may impact the current parameters of our observed laws, although such differences should be minor. Furthermore, we have conducted theoretical analyses and compared our findings with the Gaussian distribution and Zipf's law, thus dispelling any notion of coincidental discoveries in our laws.

For subsequent research, details can be drawn on the quantitative relationship between geographical entities and geographical words, as well as the quantitative driving force behind the transition from Zipf's law to our laws, as they all reflect the logic of natural language. What can be extended to is the understanding of geographic information at the sentence level. Of course, applying our laws to engineering practice is also expected and promising.

Hope that we can unravel the veil surrounding geographic information in text, not only for related engineering tasks, such as geographic information extraction, but also as a touch on knowledge boundaries.

**ACKNOWLEDEMENTS**

This work is supported by the major program of Renmin University of China (No. 21XNL019)